\documentclass[10pt,twocolumn,letterpaper]{article}

\usepackage[pagenumbers]{cvpr} 

\definecolor{cvprblue}{rgb}{0.21,0.49,0.74}
\usepackage[pagebackref,breaklinks,colorlinks,citecolor=cvprblue]{hyperref}

\usepackage{enumitem}

\usepackage{stfloats}

\usepackage[all]{nowidow}

\usepackage{microtype}

\usepackage{xcolor}

\usepackage{etoolbox}
\newcounter{BalanceAtReference}
\newcounter{ReferenceIndexForBalancing}
\makeatletter
\def\@balancelastpageonce{%
	\ifnum\value{ReferenceIndexForBalancing}=\value{BalanceAtReference}
	\newpage
	\else
	\relax
	\fi
	\stepcounter{ReferenceIndexForBalancing}
}
\pretocmd{\bibitem}{\@balancelastpageonce}
{} 
{\@latex@error{Patching \bibitem failed}{\@ehd}}
\makeatother

\def\paperID{21173} 
\def\confName{CVPR}
\def\confYear{2026}

\title{RapidPoseTriangulation: Multi-view Multi-person Whole-body \protect\\ Human Pose Triangulation in a Millisecond}

\author{Daniel Bermuth\\
University of Augsburg, Germany\\
{\tt\small daniel.bermuth@uni-a.de}
\and
Alexander Poeppel\\
University of Augsburg\\
{\tt\small poeppel@isse.de}
\and
Wolfgang Reif\\
University of Augsburg\\
{\tt\small reif@isse.de}
}

\begin{document}
\maketitle

\begin{abstract}
    The integration of multi-view imaging and pose estimation represents a significant advance in computer vision applications, offering new possibilities for understanding human movement and interactions. This work presents a new algorithm that improves multi-view multi-person pose estimation, focusing on fast triangulation speeds and good generalization capabilities. 
    The approach extends to whole-body pose estimation, capturing details from facial expressions to finger movements across multiple individuals and viewpoints. Adaptability to different settings is demonstrated through strong performance across unseen datasets and configurations. To support further progress in this field, all of this work is publicly accessible.
\end{abstract}

\vspace{-12pt}
\section{Introduction}
\label{sec:intro}
\vspace{-3pt}

\begin{figure}[htbp]
  \centering
  \begin{subfigure}{1.0\linewidth}
    \centering
    \includegraphics[width=0.87\linewidth]{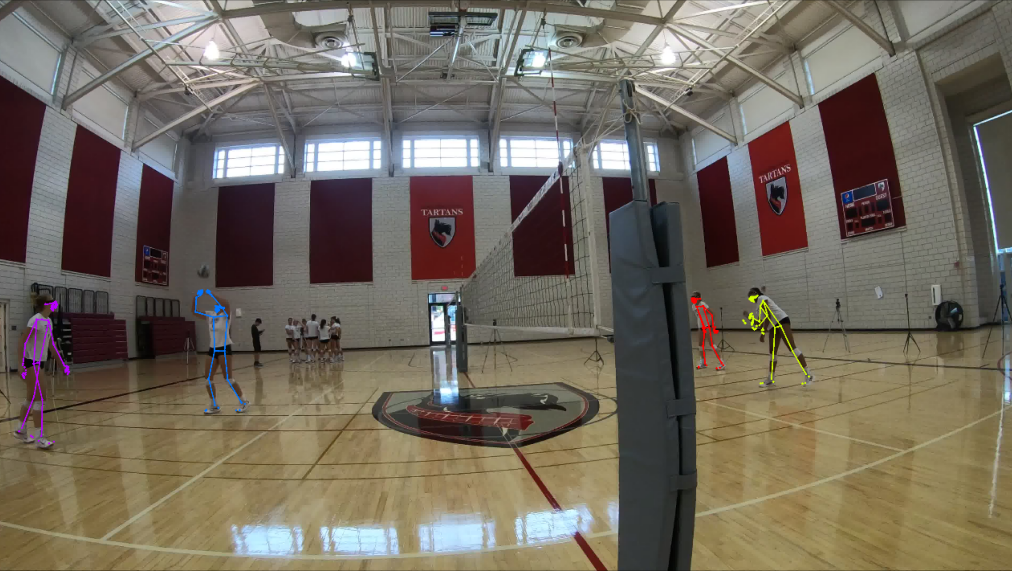}
  \end{subfigure}
  \begin{subfigure}{0.49\linewidth}
    \centering
    \includegraphics[width=0.72\linewidth]{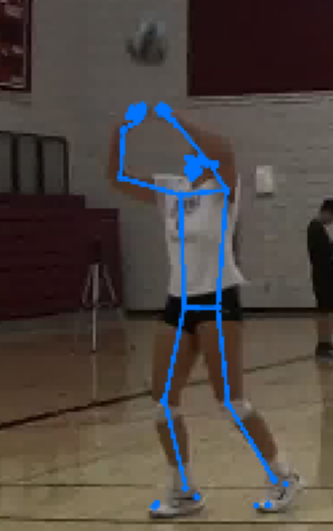}
  \end{subfigure}
  \hfill
  \begin{subfigure}{0.49\linewidth}
    \centering
    \includegraphics[width=0.72\linewidth]{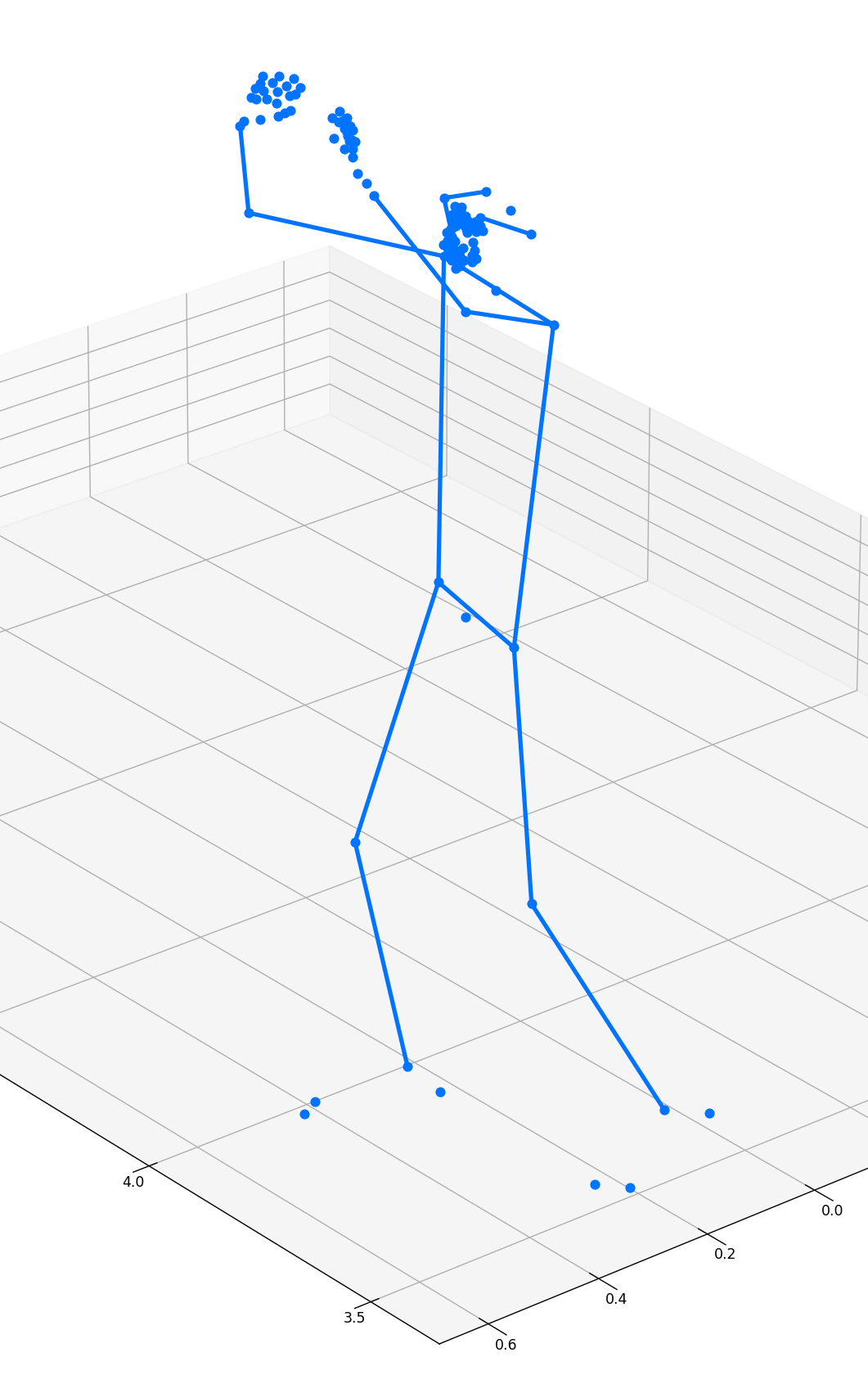}
  \end{subfigure}
  \caption{Example of a multi-person whole-body pose estimation from multiple camera views in a volleyball game (from the \textit{egohumans} dataset~\cite{khirodkar2023ego}). On top, the full image of one camera with the projections of all the detected poses, on the bottom-left a zoom-in on one player, and on the bottom-right her detected 3D pose.}
  \label{fig:init_example}
  \vspace{-6pt}
\end{figure}

The ability to accurately detect and track human body positions and movements is an important task in many advanced computer vision applications.
From enhancing virtual reality experiences to improving the collaboration with humans in robotic systems, precise human poses play a crucial role at connecting digital systems with the physical world.

Human pose estimation, which involves determining the spatial locations of different body joints, has seen significant progress in recent years.
While traditional approaches relied on wearable markers for precise tracking, the field has increasingly moved towards marker-less methods using standard cameras.
This shift offers greater flexibility and usefulness, especially in scenarios where attaching markers is impractical or impossible, such as in public spaces or specialized environments like operating rooms.

However, marker-less pose estimation, especially with only a single viewpoint, presents its own set of challenges.
Occlusions, varying lighting conditions, and the complexity of human movements can all impact the accuracy and reliability of these systems.
To address external and self-occlusion issues, and to some extent also varying lighting, multi-view approaches have become more common.
By capturing the scene from multiple angles simultaneously, these systems can overcome many of those limitations.

Recent advances in deep learning have dramatically improved the accuracy of pose estimation from individual camera views. 
However, the task of efficiently combining these multiple 2D estimates into accurate 3D poses, particularly for multiple people in real-time scenarios, remains an active area of research.
Current approaches often show notable performance drops when applied to setups they were not trained on, and are normally too slow for real-time usage~\cite{voxkeyfuse}.
Furthermore, as applications become more advanced, there is an increasing demand for a more detailed whole-body pose estimation, including facial expressions and especially hand gestures (see Figure~\ref{fig:init_example}).

\vspace{3pt}
This work addresses these challenges by introducing a novel algorithm for multi-view multi-person human pose estimation.
The approach is designed to be both fast and capable of generalizing well across different datasets and application setups.
Unlike most previous works, the algorithm is also able to estimate whole-body poses.
In a broad evaluation across multiple datasets, it is shown that the proposed algorithm is more reliable than existing methods, while also being significantly faster, which makes it especially suitable for real-time applications.

The demonstration that a lightweight algebraic method can outperform many modern learned approaches also raises an important question: Is the current trend toward increasingly complex and novelty-oriented learnable architectures truly the most effective way to improve performance in multi-view pose estimation?

To further support future research in all directions, the source-code of the presented algorithm is made available at:~\url{https://gitlab.com/Percipiote/}


\section{Related Work}
\label{sec:relwork}
\vspace{-3pt}

Most approaches address the problem of human pose estimation in two separate steps.
At first, 2D poses are predicted for each image, and in the second step, these are fused to estimate 3D poses.
The basic concepts of the different approaches in the second step can be categorized by whether they use algorithmic strategies or learning-based methods.
Some of them also make use of temporal information from previous frames to improve the results.

\vspace{3pt}
On the side of the learning-based methods, \textit{VoxelPose}~\cite{voxelpose} was one of the first concepts, building upon the voxel-based triangulation research of \textit{Iskakov et al.}~\cite{iskakov2019learnable} and extending it from single-person to multi-person estimations. 
Its method projects joint heatmaps from 2D images into a 3D voxelized space, estimates a central point for each individual using a convolution network, extracts a focused cube surrounding each person center, and computes the joint positions using a second net.
\textit{Faster-VoxelPose}~\cite{fastervoxelpose} enhanced this technique by restructuring the 3D voxel network into several 2D and 1D projections, thereby boosting efficiency.
\textit{TesseTrack}~\cite{reddy2021tessetrack} and \textit{TEMPO}~\cite{choudhury2023tempo} added a temporal dimension into the voxel space to refine poses across sequential frames.
Alternative approaches such as \textit{PRGnet}~\cite{wu2021graph} employ a graph-based methodology or directly infer 3D poses from 2D features, like in \textit{MvP}~\cite{wang2021mvp}.
\textit{SelfPose3d}~\cite{Srivastav_2024_CVPR} adopts the framework of \textit{VoxelPose} and trains both 2D and 3D networks in a self-supervised manner using randomly augmented 2D poses.

Regarding algorithmic methodologies, \textit{mvpose}~\cite{dong2019fast} solves the estimation problem in two steps.
Initially, it identifies corresponding 2D poses across images based on geometric and visual clues, and then triangulates the matching pose groups to calculate the final output.
\textit{mv3dpose}~\cite{tanke2019iterative} utilizes a graph-matching approach to assign poses through epipolar geometry and incorporates temporal information to handle missing joint data.
\textit{PartAwarePose}~\cite{chu2021part} accelerates the pose-matching process by leveraging poses from preceding frames and applying a joint-based filter to solve keypoint inaccuracies resulting from occlusions.
\textit{VoxelKeypointFusion}~\cite{voxkeyfuse} employs a voxel-based triangulation concept to predict 3D joint proposals from overlapping viewpoints.
It then uses the joint reprojections to assign them to individuals in the original 2D views, to match them to individual 3D persons before merging them into a final result.

Regarding the generalization of learning-based approaches to new datasets, a direct transfer is typically not evaluated.
However, \textit{VoxelPose}, \textit{Faster-VoxelPose}, \textit{MvP}, and \textit{TEMPO} have implemented a finetuning strategy utilizing synthetic data, though only the first two have also published the source-code for this.
The idea is to randomly place 3D poses from another dataset in 3D space, and back-project them onto the camera perspectives of the target setup.
After adding some augmentations, the network is then trained to learn the 3D reconstruction process.
In the closed-source approach of \textit{CloseMoCap}~\cite{shuai2023reconstructing} this concept was further improved by using a larger 3D pose dataset and additional augmentations.
Algorithmic approaches generally evaluated their transferability to a limited number of other datasets, a process that is considerably more straightforward as it doesn't require any training.


\section{RapidPoseTriangulation}
\label{sec:algs1}
\vspace{-3pt}

The new algorithm called \textit{RapidPoseTriangulation} follows a simple and learning-free triangulation concept. 

\subsection{Algorithm}
Similar to most other approaches, the algorithm can be split into two stages as well, with the first one predicting the 2D poses for each image.
For this any 2D pose estimator can be integrated, here \textit{RTMPose}~\cite{jiang2023rtmpose} was used.
The second stage can be split into the following steps:

\begin{enumerate}[itemsep=-1pt, topsep=3pt, partopsep=0pt, parsep=3pt, labelindent=9pt]
    \item Create all possible pairs with poses from other views
    \item Filter pairs using the previous 3D poses
    \item Select the core joints
    \item Triangulate each pair to a 3D proposal
    \item Drop proposals outside the room
    \item Reproject into 2D views
    \item Calculate reprojection error to original 2D poses
    \item Drop pairs/proposals with large errors
    \item Group the remaining 3D proposals in 3D space
    \item Triangulate again, but now with all joints
    \item Merge the proposals groups into a single 3D pose
    \item Drop clearly invalid persons
    \item Optional: Assign persons to tracks over time
    \item Optional: Clip joint movements to a maximum speed
\end{enumerate}

\begin{figure*}[htbp]
  \centering
  \begin{subfigure}{0.29\linewidth}
    \centering
    \includegraphics[width=0.71\linewidth]{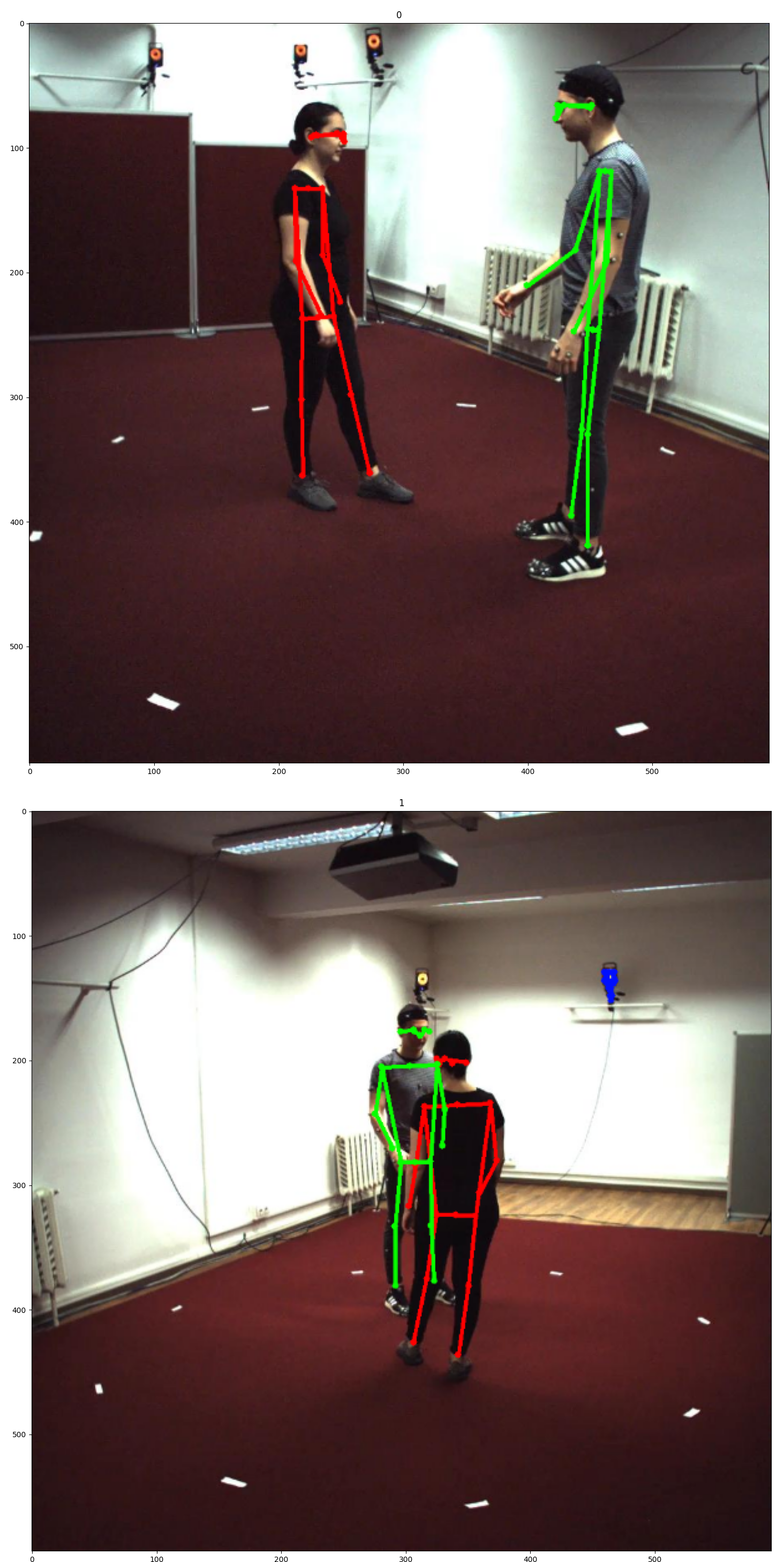}
    \caption{}
    \label{fig:vkf_personids_3}
  \end{subfigure}
  \begin{subfigure}{0.41\linewidth}
    \centering
    \includegraphics[width=0.99\linewidth]{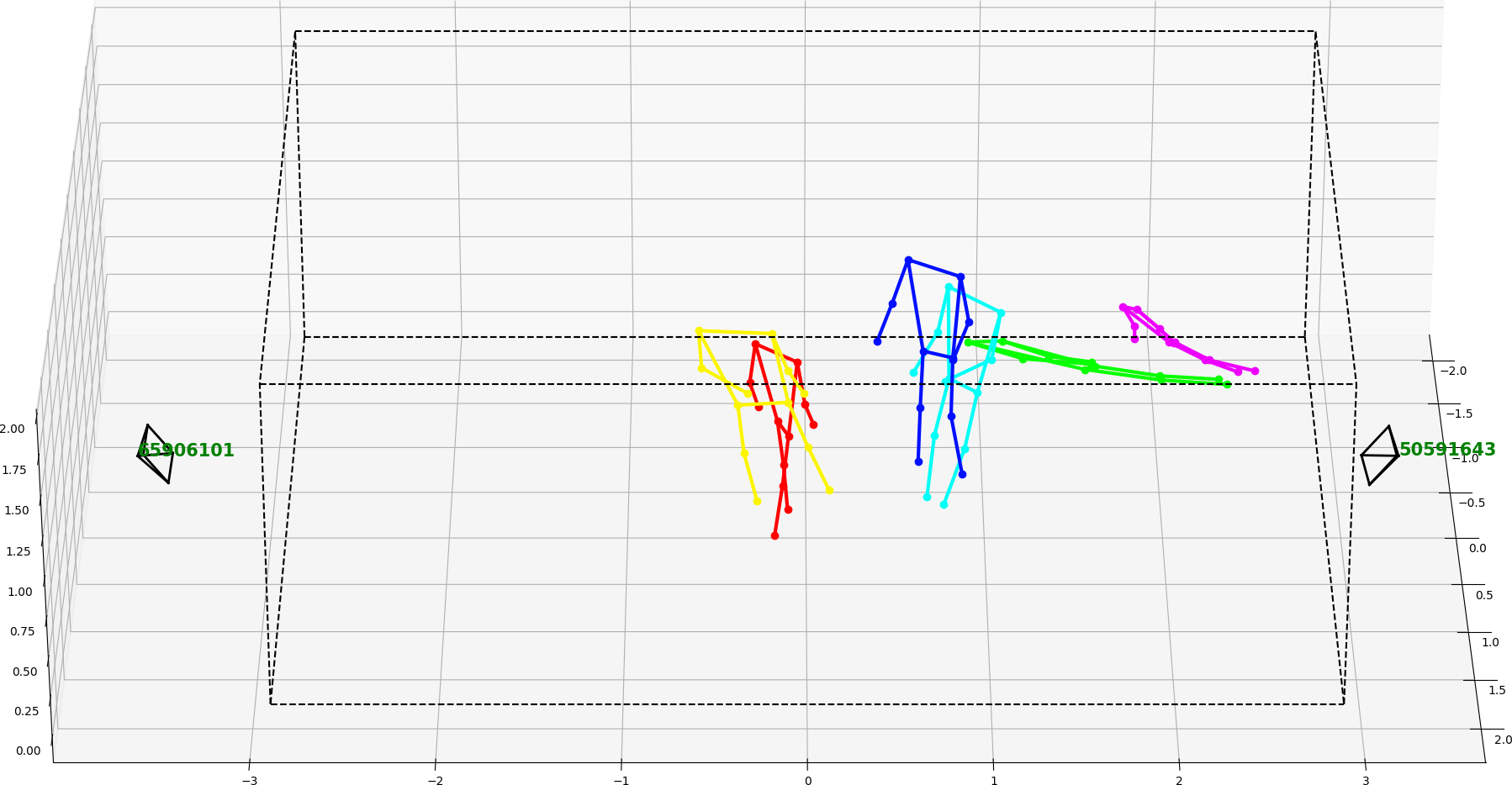}
    \hspace{12pt}
    \caption{}
    \label{fig:rpt_triang}
  \end{subfigure}
  \hfill
  \begin{subfigure}{0.29\linewidth}
    \centering
    \includegraphics[width=0.71\linewidth]{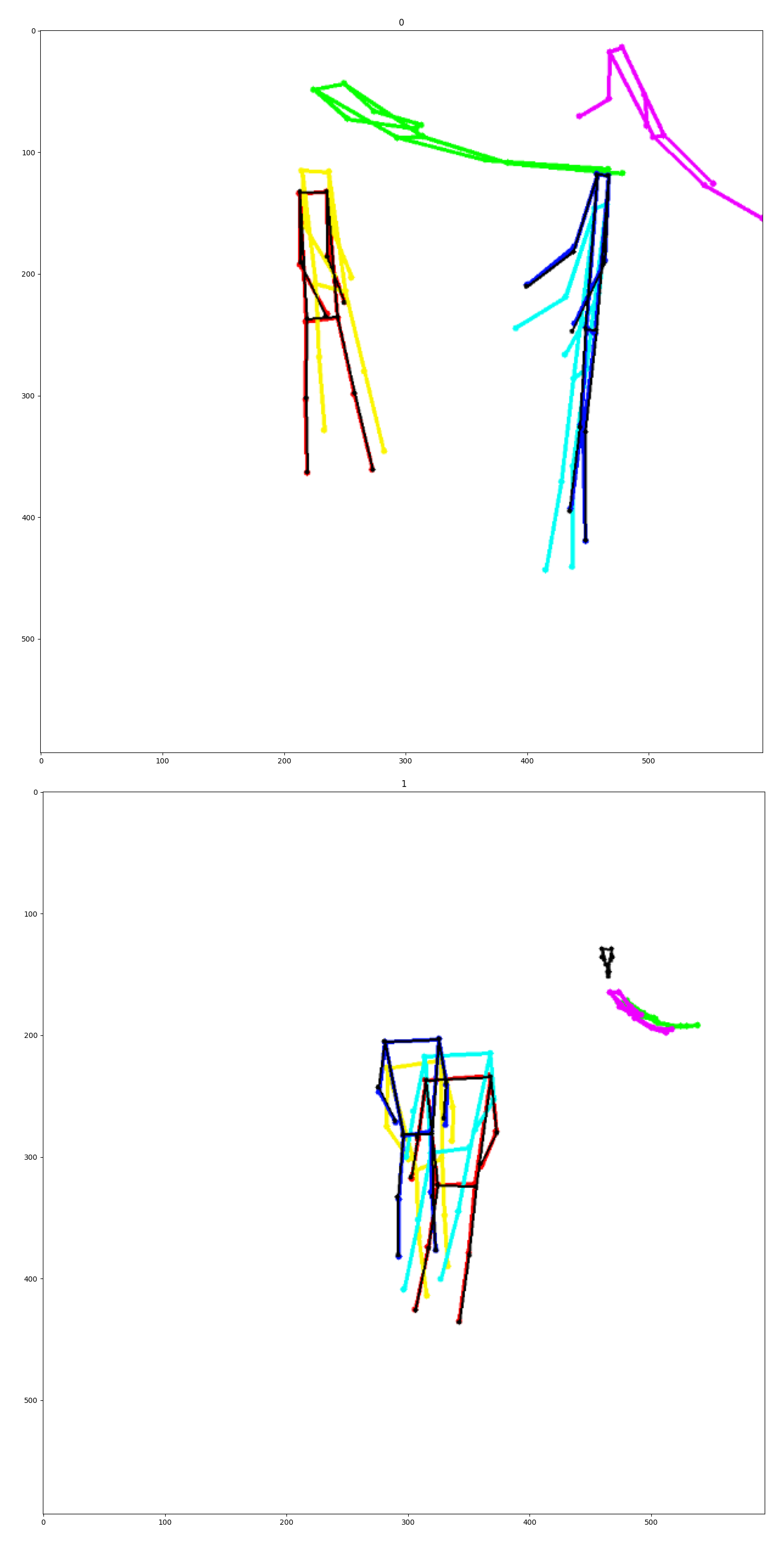}
    \caption{}
    \label{fig:vkf_personids_2}
  \end{subfigure}
  \vspace{-3pt}
  \caption{Obtaining 3D proposals. The process starts with 2D detections for each view (a, images from \textit{chi3d}~\cite{fieraru2020three}, note the small blue colored false positive in the second image). Then, in step (1), all possible pairs between the views are created. In this case this leads to six pairs from the two detections above with the three below. The core joints of all pairs are triangulated into 3D proposals (b, steps (3,4)). Then they are reprojected into the 2D views (c, step (6)), and a distance-based error to the original 2D poses (visualized in black in c) is calculated. As can be seen in the image, the green and pink proposals clearly do not match to their original 2D poses, and get a very high error. The yellow and light-blue poses resulted from the flipped (man with woman) pairs and also have a notable error. All pairs with errors above a threshold are dropped in step (8). Only the remaining red and dark-blue pairs with low enough errors are used for the further steps (9-12).}
  \label{fig:rpt_concept}
\end{figure*}

In step~(1), for each 2D pose, all possible pairs with poses from other views are created, since no association to real persons is known yet. 
In the next step~(2), the pairs are filtered using the previous 3D poses, if available.
For this, the 3D poses from the last frame are projected into the 2D views and a pixel-based distance threshold is used.
The idea is to reduce the total number of pairs that need to be triangulated.
In a pair of 2D persons in two different views, either both parts can match to one of the reprojected 3D poses, only one part is matched, or none of them.
If both pairs match, it is likely a valid pair belonging to that 3D person, and it is kept.
If only one part matches, a match is unlikely, because the 3D pose should be matchable in both views, so the second part is likely a pose from a different 3D person, and the pair is dropped.
If none is matching, it is likely that a new 3D person is present in the current frame, and the pair is kept.  
In step~(3) the core joints are extracted, which are the shoulders, hips, elbows, wrists, knees, and ankles.
They are enough to associate the following 3D proposals to a person, and this reduces the computational complexity in the next steps.
In step~(4) each pair is triangulated to a 3D proposal (see Figure~\ref{fig:rpt_triang}). 
In the following step~(5) clearly invalid proposals outside the observed room are dropped.
In step~(6) the remaining proposals are reprojected into the 2D views.
Step~(7) calculates a pixel-distance-based reprojection error to the pair's original 2D poses, and also prunes clearly invalid limbs.
In step~(8) pairs/proposals with a large error are dropped, as they are considered invalid matches.
In step~(9) the remaining 3D proposals are grouped in 3D space. Each of the 2D view pairs created a 3D proposal, and if they are close in 3D space, they likely belong to the same person, and are therefore grouped together.
Grouping in 3D makes use of physical constraints, which allows a simpler and faster implementation.
In step~(10) the remaining pairs, which normally are less than in step~(4), are triangulated again, but this time with all joints. 
In step~(11) for each group the calculated 3D proposals are merged into a single 3D pose.
To filter outliers, first for each joint an average location is calculated, and then for each proposal, the distance of this joint to the average is calculated.
If the distance is too great, the joint is dropped.
From the remaining joint proposals, the \textit{top-k} closest to the average are selected, to drop outliers again, and the final joint location is calculated as the average of those.
In general, this late fusion step allows for stronger outlier filtering, because multiple proposals can be used, but also works if a person is only visible in two views.
In step~(12) clearly invalid persons, with too few keypoints, that are too small or large, or outside the room, are dropped, and missing joints are filled with their next neighbors.
The general idea behind the algorithmic structure is to use only the minimal necessary information at each step, drop bad proposals early on, and move expensive operations, like the full-body keypoint calculations, to later steps.

Optionally, in steps~(13-14) the results can be tracked and smoothed over time.
A simple distance-based threshold is used to assign poses to existing tracks.
Just clipping the joint movements to a maximum speed was preferred to more complex smoothing filters, because it does not introduce response latency to sudden direction changes, which could hurt real-time-position accuracy. 

\subsection{Method Comparison}

There exist several differences to the other algorithmic approaches. 
In comparison to \textit{mvpose}~\cite{dong2019fast} no person identification by appearance similarities is required, the association is done only by geometric matching. 
\textit{mv3dpose}~\cite{tanke2019iterative} matches and groups 2D poses by epipolar geometry instead of the 3D triangulation and grouping used here.
\textit{PartAwarePose}~\cite{chu2021part} groups the 2D poses by assigning them to the 3D poses from the previous view, and uses a temporal filter to predict future locations from joints using previous observations to filter joints, which on the other hand can lead to prediction errors and costs additional computation time.
\textit{4DAssociation}~\cite{20204DAssociation} follows a bottom-up graph-based concept to match body-parts and connections over space and time to assemble the 3D poses afterward.
\textit{Bridgeman et al}~\cite{bridgeman2019multi} group the 2D poses to persons by first calculating correspondence costs for all pairs of 2D poses, using the minimal distance between two ray projections, calculate 3D poses and then iteratively calculate costs from the point-to-ray distance to update the 2D-3D matches.
The results are further smoothed by a temporal filter.
The source-code was not published.
\textit{QuickPose}~\cite{zhou2022quickpose} splits the detected 2D skeletons into all plausible part-graphs, calculates a ray-based distance between joints in different views, and uses a clustering algorithm to find matching parts.
\textit{Chen et al.}~\cite{Chen_2020_CVPR} iteratively match 2D poses from one view to the already existing 3D persons, and gradually update the matched 3D poses, before continuing with the next view.
The last two algorithms are the fastest ones yet, according to the times stated in their papers, but their source-codes were not published, therefore a detailed comparison is not possible.
In comparison to \textit{VoxelKeypointFusion}~\cite{voxkeyfuse} no voxel space is used, which results in much faster calculations, and its person association follows a bottom-up instead of the top-down approach used here.

\subsection{Performance}

Due to the relatively simple concept, the algorithm is very fast.
In a test on the \textit{shelf}~\cite{belagiannis20143d} dataset, which shows up to four persons in five views, the algorithm requires an average triangulation time of only $0.1\,ms$.
The per-step times are listed in Table~\ref{tab:steps}. 

\begin{table}[htbp]
  \fontsize{7pt}{7pt}\selectfont
  \centering
  \begin{tabular}{@{}|l|c|@{}}
    \toprule
    Method \hspace{44pt}                        & Time\,(µs) \\
    \midrule
    Pose undistortion                           & 14         \\
    Pair creation (1)                           & 4          \\
    Pair filtering (2)                          & 10          \\
    Triangulate and Score (3-7)                 & 28         \\
    Grouping (9)                                & 7          \\
    Triangulate and Score (10)                  & 28         \\
    Merge (11)                                  & 12         \\
    Post-process (12)                           & 7          \\
    Tracking (13-14)                            & 1          \\
    \midrule
    Total                                       & 115        \\   
    \bottomrule
  \end{tabular}
  \caption{Average time for different steps in microseconds.}
  \label{tab:steps}
\end{table}

While the initial version of the algorithm used \textit{OpenCV} functions for most calculations, the current version uses standard \textit{C++} functions only, because the overhead of the \textit{OpenCV} functions was higher than the actual calculation time for the relatively small number of points.
Instead of \textit{OpenCV's} \textit{SVD}-based triangulation, a simple mid-point triangulation was implemented, which was much faster and almost as accurate in the experiments.
In summary the switch to standard \textit{C++} functions led to a significant relative speedup of around $0.1\,ms$.
Using only the core joints, and repeatedly filtering invalid poses helps to keep the subsequent steps efficient.
The runtime is further influenced by the number of joints, which are by default~$20$.
In the whole-body case that uses $136$ joints, the average runtime increases to $0.4\,ms$.
Depending on the integration of the algorithm, data conversion time can lead to further delays.
The algorithm itself is implemented in \textit{C++}, but an interface to \textit{Python} (using \textit{SWIG}) is provided as well.

\vspace{3pt}
As shown in Table~\ref{tab:times} the proposed solution is multiple times faster than all the other algorithms, which proves its significant efficiency improvements.
Note that all times reported in the different papers were measured on different hardware (see appendix), here a single core of an \textit{AMD-7900X} was used, but the main speedup is clearly caused by the more efficient algorithm.

\begin{table}[htbp]
  \fontsize{7pt}{7pt}\selectfont
  \centering
  \begin{tabular}{@{}|l|c|@{}}
    \toprule
    Method \hspace{44pt}                        & Time\,(ms) \\
    \midrule
    mvpose~\cite{dong2019fast}                  & 105  \\
    VoxelPose                                   & 103  \\
    mv3dpose                                    & 63   \\
    PlaneSweepPose                              & 57   \\
    VoxelKeypointFusion~\cite{voxkeyfuse}       & 48   \\
    Faster-VoxelPose                            & 38   \\
    4DAssociation~\cite{zhou2022quickpose}      & 32   \\
    PartAwarePose~\cite{chu2021part}            & 10   \\
    Bridgeman et al.~\cite{bridgeman2019multi}  & 9.1  \\
    Chen et al.~\cite{Chen_2020_CVPR}           & 3.1  \\
    QuickPose~\cite{zhou2022quickpose}          & 2.9  \\
    \midrule
    RapidPoseTriangulation                      & 0.1  \\   
    \bottomrule
  \end{tabular}
  \caption{Time from 2D to 3D poses on \textit{shelf}~\cite{belagiannis20143d} in milliseconds.}
  \label{tab:times}
\end{table}

\vspace{-3pt}
\section{Dataset Generalization}
\label{sec:ndgen1}
\vspace{-3pt}

The most important part for machine learning models is their generalization to previously unknown setups.
Therefore the proposed approach was evaluated on a variety of datasets.
The datasets used for the evaluation were \textit{human36m}~\cite{h36m_pami}, \textit{shelf}~\cite{belagiannis20143d}, \textit{campus}~\cite{belagiannis20143d}, \textit{mvor}~\cite{srivastav2018mvor}, and \textit{panoptic}~\cite{joo2015panoptic}, similar as in other works. 
Additionally, the datasets \textit{chi3d}~\cite{fieraru2020three}, \textit{tsinghua}~\cite{20204DAssociation} and \textit{egohumans}~\cite{khirodkar2023ego} were added as well to allow an even broader evaluation.
The metrics used are the same as in \textit{VoxelKeypointFusion}~\cite{voxkeyfuse}, and evaluate the same $13$ keypoints (2~shoulders, 2~hips, 2~elbows, 2~wrists, 2~knees, 2~ankles, 1~nose/head) across each dataset.
The metrics follow common standards and are described in more detail in~\cite{voxkeyfuse}.
The \textit{FPS} were measured on an \textit{Nvidia-3090} as well.
Due to brevity and readability reasons, for the datasets already evaluated in \textit{VoxelKeypointFusion}~\cite{voxkeyfuse}, only the best results/algorithms are compared here.

\begin{table*}[htbp]
    \fontsize{7pt}{7pt}\selectfont
    \centering
    \begin{tabular}{@{}|l|c|cc|c|cc|c|c|c|c|@{}}
      \toprule
      Method \hspace{44pt}          & {\,}PCP{\,}   & \multicolumn{2}{c|}{PCK@100/500}          & {\,}MPJPE{\,} & \multicolumn{2}{c|}{Recall@100/500}       & {\,}Invalid{\,} & {}{\,}F1{\,}{} & {\,}FPS{\,}   \\
      \midrule
      Faster-VoxelPose\,(synthetic) & 91.3          & \hspace{6pt} 75.5          & 98.8         & 88.3          & \hspace{6pt} 75.0          & \textbf{100} & 0.2             & 99.5           & 36.2          \\
      VoxelKeypointFusion           & \textbf{96.9} & \hspace{6pt} 81.7          & \textbf{100} & 64.3          & \hspace{6pt} \textbf{95.0} & \textbf{100} & \textbf{0}      & \textbf{100}   & 8.7           \\
      \midrule
      RapidPoseTriangulation        & 96.8          & \hspace{6pt} \textbf{88.2} & 99.9         & \textbf{60.6} & \hspace{6pt} 94.5          & \textbf{100} & \textbf{0}      & \textbf{100}   & \textbf{115}  \\
      \bottomrule
    \end{tabular}
    \caption{Transfer to \textit{human36m}~\cite{h36m_pami}, all other results from \cite{voxkeyfuse}.}
    \label{tab:trans_h36m}
\vspace{12pt}
    \fontsize{7pt}{7pt}\selectfont
    \centering
    \begin{tabular}{@{}|l|c|cc|c|cc|c|c|c|@{}}
      \toprule
      Method \hspace{44pt}          & {\,}PCP{\,}   & \multicolumn{2}{c|}{PCK@100/500}          & {\,}MPJPE{\,} & \multicolumn{2}{c|}{Recall@100/500}       & {\,}Invalid{\,} & {}{\,}F1{\,}{} & {\,}FPS{\,}   \\
      \midrule
      Faster-VoxelPose              & 99.1          & \hspace{6pt} 88.3          & \textbf{100} & 59.8          & \hspace{6pt} \textbf{99.4} & \textbf{100} & 50.7            & 66.0           & 17.6          \\
      Faster-VoxelPose\,(synthetic) & 99.0          & \hspace{6pt} 87.9          & \textbf{100} & 58.8          & \hspace{6pt} 99.0          & \textbf{100} & 48.8            & 67.7           & 17.4          \\
      MvP\,(synthetic)              & 98.6          & \hspace{6pt} 94.1          & 99.7         & 51.8          & \hspace{6pt} 97.1          & \textbf{100} & 82.2            & 30.2           & 8.5           \\
      mv3dpose                      & 97.1          & \hspace{6pt} 91.4          & 98.4         & 55.8          & \hspace{6pt} 94.8          & 98.5         & 44.3            & 71.2           & 1.6           \\
      VoxelKeypointFusion           & 98.8          & \hspace{6pt} 93.3          & \textbf{100} & 51.3          & \hspace{6pt} 98.3          & \textbf{100} & 49.1            & 67.4           & 5.8           \\
      \midrule
      RapidPoseTriangulation        & \textbf{99.2} & \hspace{6pt} \textbf{94.4} & \textbf{100} & \textbf{47.5} & \hspace{6pt} 98.7          & \textbf{100} & \textbf{41.3}   & \textbf{74.0}  & \textbf{66.6} \\
      \bottomrule
    \end{tabular}
    \caption{Transfer to \textit{shelf}~\cite{belagiannis20143d}, all other results from \cite{voxkeyfuse}.}
    \label{tab:trans_shelf}
\vspace{12pt}
    \fontsize{7pt}{7pt}\selectfont
    \centering
    \begin{tabular}{@{}|l|c|cc|c|cc|c|c|c|c|@{}}
      \toprule
      Method \hspace{44pt}          & {\,}PCP{\,}   & \multicolumn{2}{c|}{PCK@100/500}           & {\,}MPJPE{\,} & \multicolumn{2}{c|}{Recall@100/500}       & {\,}Invalid{\,} & {}{\,}F1{\,}{} & {\,}FPS{\,}   \\
      \midrule
      Faster-VoxelPose\,(synthetic) & 65.9          & \hspace{6pt} 41.3          & 74.4          & 104           & \hspace{6pt} 39.6          & 74.7         & \textbf{2.1}    & 84.8           & 25.3          \\
      mvpose                        & 91.3          & \hspace{6pt} 70.2          & 99.9          & 80.4          & \hspace{6pt} 82.7          & \textbf{100} & 25.4            & 85.5           & 0.5           \\
      mv3dpose                      & 84.1          & \hspace{6pt} 64.4          & 93.4          & 135           & \hspace{6pt} 62.5          & 94.9         & 10.1            & \textbf{92.4}  & 2.8           \\
      PartAwarePose                 & 93.2          & \hspace{6pt} 78.4          & 98.9          & \textbf{74.7} & \hspace{6pt} \textbf{93.9} & 98.9         & 22.7            & 86.8           & 5.8           \\
      VoxelKeypointFusion           & 91.1          & \hspace{6pt} 74.6          & 99.9          & 84.4          & \hspace{6pt} 80.3          & \textbf{100} & 35.5            & 78.4           & 7.8           \\
      \midrule
      RapidPoseTriangulation        & \textbf{95.2} & \hspace{6pt} \textbf{79.9} & \textbf{100}  & 75.2          & \hspace{6pt} \textbf{93.9} & \textbf{100} & 15.9            & 91.4           & \textbf{142}  \\
      \bottomrule
    \end{tabular}
    \caption{Transfer to \textit{campus}~\cite{belagiannis20143d}, all other results from \cite{voxkeyfuse}.}
    \label{tab:trans_campus}
\vspace{12pt}
    \fontsize{7pt}{7pt}\selectfont
    \centering
    \begin{tabular}{@{}|l|c|cc|c|cc|c|c|c|c|@{}}
      \toprule
      Method \hspace{44pt}                                    & {\,}PCP{\,}   & \multicolumn{2}{c|}{PCK@100/500}           & {\,}MPJPE{\,} & \multicolumn{2}{c|}{Recall@100/500}        & {\,}Invalid{\,} & {}{\,}F1{\,}{} & {\,}FPS{\,}   \\
      \midrule
      SimpleDepthPose~\cite{simdepose}                        & \textbf{74.0} & \hspace{6pt} \textbf{62.0} & \textbf{94.3} & 113           & \hspace{6pt} \textbf{54.1} & \textbf{96.6} & 23.5            & \textbf{85.4}  & \textbf{37.2} \\
      \midrule
      \midrule
      Faster-VoxelPose\,(synthetic)~\cite{voxkeyfuse}         & 37.9          & \hspace{6pt} 28.1          & 45.5          & \textbf{109}  & \hspace{6pt} 29.4          & 46.1          & \textbf{7.7}    & 61.5           & 30.0          \\
      VoxelKeypointFusion~\cite{voxkeyfuse}                   & 54.5          & \hspace{6pt} 43.9          & 75.1          & 128           & \hspace{6pt} 35.9          & 76.6          & 24.2            & 76.2           & 11.3          \\
      \midrule
      RapidPoseTriangulation                                  & \textbf{59.0} & \hspace{6pt} \textbf{44.6} & \textbf{76.7} & 120           & \hspace{6pt} \textbf{39.9} & \textbf{77.9} & 22.1            & \textbf{77.9}  & \textbf{121}  \\
      \bottomrule
    \end{tabular}
    \caption{Transfer to \textit{mvor}~\cite{srivastav2018mvor} with depth images (first) and without (others).}
    \label{tab:trans_mvor}
\vspace{12pt}
    \fontsize{7pt}{7pt}\selectfont
    \centering
    \begin{tabular}{@{}|l|c|cc|c|cc|c|c|c|c|@{}}
      \toprule
      Method \hspace{56pt}   & {\,}PCP{\,}   & \multicolumn{2}{c|}{PCK@100/500}           & {\,}MPJPE{\,} & \multicolumn{2}{c|}{Recall@100/500}        & {\,}Invalid{\,} & {}{\,}F1{\,}{} & {\,}FPS{\,}   \\
      \midrule
      Faster-VoxelPose       & 99.4          & \hspace{6pt} 98.6          & \textbf{99.9} & 20.5          & \hspace{6pt} 99.7          & \textbf{99.9} & \textbf{1.0}    & \textbf{99.5}  & 18.0          \\
      PRGnet                 & \textbf{99.5} & \hspace{6pt} \textbf{99.1} & \textbf{99.9} & 17.1          & \hspace{6pt} \textbf{99.9} & \textbf{99.9} & 2.0             & 99.0           & 6.8           \\
      TEMPO                  & 98.1          & \hspace{6pt} 97.4          & 98.5          & \textbf{16.8} & \hspace{6pt} 98.4          & 98.4          & 2.4             & 98.0           & 5.1           \\
      \midrule
      \midrule
      VoxelKeypointFusion    & 97.1          & \hspace{6pt} 94.0          & 99.7          & 47.8          & \hspace{6pt} 97.3          & \textbf{99.9} & 2.4             & 98.7           & 4.2           \\
      \midrule
      RapidPoseTriangulation & \textbf{99.1} & \hspace{6pt} \textbf{96.5} & \textbf{99.8} & \textbf{30.6} & \hspace{6pt} \textbf{99.3} & 99.8          & 1.8             & \textbf{99.0}  & \textbf{57.5} \\
      \bottomrule
    \end{tabular}
    \caption{Replication of \textit{panoptic}~\cite{joo2015panoptic} results and transfer without depth, all other results from \cite{voxkeyfuse}.}
    \label{tab:res_panoptic}
\end{table*}

\subsection{Standard datasets}

On the commonly used datasets, \textit{RapidPoseTriangulation} shows a similar or better generalization performance than most state-of-the-art algorithms, while reaching an outstanding speed at the same time.
See Tables~\ref{tab:trans_h36m},\,\ref{tab:trans_shelf},\,\ref{tab:trans_campus},\,\ref{tab:trans_mvor},\,\ref{tab:res_panoptic}.

Much of the \textit{MPJPE} error in the \textit{human36m} dataset results from different skeleton definitions.
The original labels for example have higher hips, the nose more inside the head, and the ankles more at the heels, than the \textit{COCO}~\cite{lin2014microsoft} skeleton definition of the 2D pose estimation network.
From a human perspective though, the predictions would often match better to the images than the original labels. 

Regarding the \textit{campus\,\&\,shelf} datasets, a visual inspection of samples with large errors showed that a majority of them are caused by label errors, and not by the algorithm itself.
Therefore it needs to be noted that the comparability of the results of well-performing algorithms seems to be somewhat limited.

On \textit{mvor} the algorithm suffers, similar to the other RGB-only approaches, from high occlusions and few viewpoints. 

Many errors in the \textit{panoptic} dataset involve the lower body joints, especially the ankles.
While most upper body joints have an average error between \mbox{$15$\textit{-}$25\,mm$}, the hips and knees have around \mbox{$35$\textit{-}$40\,mm$}, and the ankles around $50\,mm$.
This is mainly caused by the position of the cameras, which often cut off part of the legs.
The 2D pose estimation network, however, still predicts those joints at the lower edge of the image (so for example an ankle is predicted at the height of the knee).
This leads to a large error in the triangulation step because the 3D joint is then estimated with a wrong depth, which cannot always be detected as an outlier.
For future optimization, this might be better handled if the 2D pose estimation network could predict the visibility of the keypoints as well, so the algorithm could prefer directly visible keypoints over estimated ones.

\subsection{Additional datasets}

To allow an even broader comparison, two further datasets were added, following the same scheme as in \textit{VoxelKeypointFusion}~\cite{voxkeyfuse}, and using their \textit{skelda} dataset library.

\begin{table*}[htbp]
  \fontsize{7pt}{7pt}\selectfont
  \centering
  \begin{tabular}{@{}|l|c|cc|c|cc|c|c|c|c|@{}}
    \toprule
    Method \hspace{44pt}          & {\,}PCP{\,}   & \multicolumn{2}{c|}{PCK@100/500}           & {\,}MPJPE{\,} & \multicolumn{2}{c|}{Recall@100/500}        & {\,}Invalid{\,} & {}{\,}F1{\,}{} & {\,}FPS{\,}   \\
    \midrule
    VoxelPose\,(synthetic)        & 97.4          & \hspace{6pt} 88.4          & 99.4          & 72.2          & \hspace{6pt} 90.5          & \textbf{100}  & 0.5             & 99.7           & 15.2          \\
    Faster-VoxelPose\,(synthetic) & 97.3          & \hspace{6pt} 88.2          & 99.2          & 69.3          & \hspace{6pt} 94.9          & 99.6          & 0.8             & 99.4           & 31.9          \\
    PRGnet                        & 95.8          & \hspace{6pt} 82.9          & 99.1          & 80.1          & \hspace{6pt} 85.0          & \textbf{100}  & 10.2            & 94.6           & 5.0           \\
    TEMPO                         & 89.0          & \hspace{6pt} 81.1          & 90.7          & 66.5          & \hspace{6pt} 86.4          & 91.0          & 6.6             & 92.2           & 8.9           \\
    SelfPose3d                    & 79.7          & \hspace{6pt} 71.5          & 87.2          & 97.8          & \hspace{6pt} 66.5          & 89.7          & 79.8            & 32.9           & 5.1           \\
    mvpose                        & 98.5          & \hspace{6pt} 85.2          & \textbf{100}  & 66.7          & \hspace{6pt} 98.3          & \textbf{100}  & 0.5             & 99.7           & 0.2           \\
    mv3dpose                      & 58.9          & \hspace{6pt} 53.8          & 61.1          & 71.8          & \hspace{6pt} 55.9          & 61.4          & 2.2             & 75.4           & 2.6           \\
    PartAwarePose                 & 89.1          & \hspace{6pt} 81.6          & 92.9          & 77.8          & \hspace{6pt} 84.0          & 94.0          & 8.7             & 92.5           & 3.4           \\
    VoxelKeypointFusion           & 94.8          & \hspace{6pt} 81.7          & 99.1          & 68.5          & \hspace{6pt} 95.5          & 99.4          & 0.1             & 99.6           & 8.0           \\
    \midrule
    RapidPoseTriangulation        & \textbf{99.0} & \hspace{6pt} \textbf{90.5} & 99.9          & \textbf{60.3} & \hspace{6pt} \textbf{98.6} & \textbf{100}  & \textbf{0}      & \textbf{100}   & \textbf{97.3} \\
    \bottomrule
  \end{tabular}
  \caption{Transfer to \textit{chi3d}~\cite{fieraru2020three}}
  \label{tab:trans_chi3d}
\end{table*}

\begin{table*}[htbp]
  \fontsize{7pt}{7pt}\selectfont
  \centering
  \begin{tabular}{@{}|l|c|cc|c|cc|c|c|c|c|@{}}
    \toprule
    Method \hspace{44pt}          & {\,}PCP{\,}   & \multicolumn{2}{c|}{PCK@100/500}           & {\,}MPJPE{\,} & \multicolumn{2}{c|}{Recall@100/500}        & {\,}Invalid{\,} & {}{\,}F1{\,}{} & {\,}FPS{\,}   \\
    \midrule
    VoxelPose\,(synthetic)        & 97.8          & \hspace{6pt} 94.1          & 99.1          & 58.9          & \hspace{6pt} 94.1          & 99.5          & 5.7             & 96.8           & 5.1           \\
    Faster-VoxelPose\,(synthetic) & 94.8          & \hspace{6pt} 87.9          & 98.6          & 66.3          & \hspace{6pt} 89.7          & 99.0          & 9.2             & 94.8           & 7.7           \\
    PRGnet                        & 96.8          & \hspace{6pt} 88.7          & 99.9          & 67.8          & \hspace{6pt} 93.2          & \textbf{100}  & 10.5            & 94.5           & 6.3           \\
    TEMPO                         & 89.2          & \hspace{6pt} 85.7          & 90.4          & 57.3          & \hspace{6pt} 89.4          & 90.3          & 1.4             & 94.3           & 5.0           \\
    SelfPose3d                    & 87.2          & \hspace{6pt} 83.1          & 89.2          & 65.9          & \hspace{6pt} 83.0          & 90.2          & 51.2            & 63.3           & 4.9           \\
    mvpose                        & 91.4          & \hspace{6pt} 81.5          & 99.3          & 79.2          & \hspace{6pt} 88.6          & 99.7          & 4.7             & 97.4           & 0.4           \\
    mv3dpose                      & 68.6          & \hspace{6pt} 64.8          & 71.0          & 63.3          & \hspace{6pt} 66.8          & 71.3          & 14.6            & 77.7           & 1.2           \\
    PartAwarePose                 & 89.9          & \hspace{6pt} 81.7          & 94.3          & 79.7          & \hspace{6pt} 78.6          & 94.9          & 6.7             & 94.1           & 0.8           \\
    VoxelKeypointFusion           & 98.0          & \hspace{6pt} 95.9          & 99.5          & \textbf{51.1} & \hspace{6pt} 95.1          & 99.8          & 0.9             & 99.4           & 3.1           \\
    \midrule
    RapidPoseTriangulation        & \textbf{98.7} & \hspace{6pt} \textbf{96.0} & \textbf{99.8} & 52.8          & \hspace{6pt} \textbf{95.7} & \textbf{100}  & \textbf{0}      & \textbf{100}   & \textbf{48.4} \\
    \bottomrule
  \end{tabular}
  \caption{Transfer to \textit{tsinghua}~\cite{20204DAssociation}}
  \vspace{-6pt}
  \label{tab:trans_tsinghua}
\end{table*}

\begin{table*}[!hbp]
  \vspace{-6pt}
  \fontsize{7pt}{7pt}\selectfont
  \centering
  \begin{tabular}{@{}|l|c|cc|c|cc|c|c|c|@{}}
      \toprule
  
  Method \hspace{44pt} & {\,}PCP{\,}   & \multicolumn{2}{c|}{PCK@100/500}           & {\,}MPJPE{\,} & \multicolumn{2}{c|}{Recall@100/500}       & {\,}Invalid{\,} & {}{\,}F1{\,}{} & {\,}FPS{\,}   \\
  \midrule

  VoxelPose\,(synthetic) & 81.4 & 68.7 & 90.5 & 100 & 64.6 & 92.1 & 26.7 & 79.3 & 10.7 \\
  Faster-VoxelPose\,(synthetic) & 81.0 & 68.1 & 86.1 & 82.6 & 71.3 & 86.6 & \textbf{11.5} & 84.6 & 24.8 \\
  PRGnet & 63.5 & 52.4 & 68.4 & 134 & 51.3 & 69.3 & 28.8 & 63.3 & 9.9 \\
  TEMPO & 69.1 & 59.9 & 72.2 & 79.2 & 61.9 & 72.4 & 21.2 & 69.7 & 9.9 \\
  SelfPose3d & 68.0 & 55.6 & 77.4 & 115 & 52.4 & 79.0 & 68.3 & 42.5 & 7.0 \\
  mvpose & 83.4 & 70.5 & 90.0 & 81.6 & 76.8 & 90.4 & 19.6 & 83.3 & 0.4 \\
  mv3dpose & 61.0 & 53.9 & 64.4 & 110 & 54.6 & 65.0 & 21.5 & 66.4 & 2.3 \\
  PartAwarePose & 79.1 & 71.0 & 83.9 & 91.5 & 74.0 & 84.6 & 17.7 & 79.9 & 3.9 \\
  VoxelKeypointFusion & 89.0 & 78.5 & 95.6 & 74.6 & 83.4 & 96.0 & 18.3 & 86.8 & 7.5 \\
  \midrule
  RapidPoseTriangulation & \textbf{91.3} & \textbf{82.3} & \textbf{96.0} & \textbf{69.4} & \textbf{86.9} & \textbf{96.3} & 13.2 & \textbf{90.5} & \textbf{98.4} \\
  \bottomrule
\end{tabular}
\caption{Averaged generalization on all six unseen datasets (\textit{human36m}, \textit{shelf}, \textit{campus}, \textit{mvor}, \textit{chi3d}, \textit{tsinghua}).}
\label{tab:trans_avg6}
\end{table*}

\vspace{3pt}
The first one is the \textit{chi3d}~\cite{fieraru2020three} dataset, which shows two persons interacting with each other.
It has the challenge that the persons are often very close to each other, which makes it hard to distinguish between them.
The setup is similar to \textit{human36m} and also contains four cameras.
The results in Table~\ref{tab:trans_chi3d} show that generally most persons are detected, with a similar performance as in \textit{human36m}.
Regarding \textit{RapidPoseTriangulation}, it could be seen that in some cases the 2D pose estimation struggled to correctly detect the joints between very close persons, and the 3D triangulation then did not get enough inputs to correctly triangulate all joints and could miss a person.
One could reduce the reprojection matching threshold to better handle such problematic 2D detections, but at the cost of increasing the number of false positives, or alternatively use the tracking information to fill-up missed persons with their last positions.
The authors of \textit{CloseMoCap}~\cite{shuai2023reconstructing} also reported very good transfer results on this dataset, reaching a \textit{PCK@50}~of~$94.3$ on average.
They also evaluated \textit{VoxelPose}, \textit{Faster-VoxelPose} and \textit{mvpose} on the same dataset and reached much better results for them as well. 
Additionally, they reported per-joint errors on a subset of the dataset and found that the hip joints had the best results, whereas here the hip joints are among the worst results.
In the labels, however, it can be seen that the hips are positioned higher on the body than in the \textit{COCO} skeleton definition, similar as already in \textit{human36m}.
So in conclusion, the mismatch is likely caused by parts of the evaluation, or the use of different labels, somewhere in \textit{CloseMoCap}, but because their source-code was not published, this assumption cannot be verified.

\vspace{3pt}
The second dataset is \textit{tsinghua}~\cite{20204DAssociation}, which shows people moving around in a room, while they are being recorded from six different cameras. 
To prevent multiple trainings for the synthetic models, only the first of the two sequences was evaluated.
The results in Table~\ref{tab:trans_tsinghua} show that almost all models perform relatively well on this dataset, but still many of them miss a few persons. 

\vspace{3pt}
Under the assumption that the general goal of all developed pose estimators is their usage in practical applications, their generalization to unseen datasets is very important. 
As can be seen in the experiments above, many models have setups in which they show good, and others in which they show bad performance. 
For a simpler algorithm comparison, the average performance on all six datasets was calculated.
As can be easily seen in Table~\ref{tab:trans_avg6}, \textit{RapidPoseTriangulation} shows the overall best performance.

\subsection{Using more cameras}
\vspace{-3pt}

The previous results demonstrate that the number of cameras has a large impact on the performance of the algorithms.
Especially \textit{mvor} shows that using only three cameras in the presence of strong occlusions leads to a large decrease in performance.
From an application point of view, it is therefore recommended to use at least five or even more cameras, if this is possible.

To evaluate the impact of a high number of cameras, the \textit{egohumans}~\cite{khirodkar2023ego}  dataset was selected.
It shows between 2-8 people in different indoor and outdoor settings (like building with lego bricks, fencing, or playing tag, basketball, volleyball, badminton, or tennis). 
They are watched by 8-20 cameras.
The supervised space is up to $25\times14\,m$ in size, which is much larger compared to the other datasets.
The last recording of each setting was selected as test set, and each recording was subsampled into parts of $3s$ consecutive motions with larger gaps in between as well.

\begin{table*}[htbp]
  \fontsize{7pt}{7pt}\selectfont
  \centering
  \begin{tabular}{@{}|l|l|c|cc|c|cc|c|c|ccc|@{}}
    \toprule
    Subset       & Method \hspace{44pt}          & {\,}PCP{\,}   & \multicolumn{2}{c|}{PCK@100/500}           & {\,}MPJPE{\,} & \multicolumn{2}{c|}{Recall@100/500}        & {\,}Invalid{\,} & {}{\,}F1{\,}{} & \multicolumn{3}{c|}{Time: Demosaic-2D-3D}        \\
    \midrule
    legoassemble & VoxelPose\,(synthetic)        & 89.9          & \hspace{6pt} 74.4          & 93.8          & 94.7          & \hspace{6pt} 65.3          & 96.1          & 71.2            & 44.4           & \hspace{6pt} & & \hspace{-26pt} 392              \\
                 & Faster-VoxelPose\,(synthetic) & 75.4          & \hspace{6pt} 64.1          & 94.2          & 107           & \hspace{6pt} 54.0          & 95.3          & \textbf{0}      & 97.6           & \hspace{6pt} & & \hspace{-26pt} 239              \\
                 & VoxelKeypointFusion           & 99.4          & \hspace{6pt} 97.2          & 99.4          & 37.0          & \hspace{6pt} 99.4          & 99.7          & \textbf{0}      & 99.9           & \hspace{6pt} & 205 & 271                         \\
                 & RapidPoseTriangulation        & \textbf{100}  & \hspace{6pt} \textbf{99.0} & \textbf{100}  & \textbf{25.3} & \hspace{6pt} \textbf{100}  & \textbf{100}  & \textbf{0}      & \textbf{100}   & \hspace{6pt} 11.2 & \textbf{23.3} & \textbf{0.3} \\
    \midrule
    tennis       & VoxelPose\,(synthetic)        & 66.9          & \hspace{6pt} 29.5          & 86.5          & 128           & \hspace{6pt} 3.9           & 87.0          & 1.3             & 92.5           & \hspace{6pt} & & \hspace{-26pt} 1077             \\
                 & VoxelKeypointFusion           & 98.2          & \hspace{6pt} 90.6          & 100           & 64.6          & \hspace{6pt} 98.1          & \textbf{100}  & \textbf{0}      & \textbf{100}   & \hspace{6pt} & 434 & 679                         \\
                 & RapidPoseTriangulation        & \textbf{99.9} & \hspace{6pt} \textbf{99.5} & \textbf{100}  & \textbf{22.7} & \hspace{6pt} \textbf{99.7} & \textbf{100}  & \textbf{0}      & \textbf{100}   & \hspace{6pt} 26.8 & \textbf{50.1} & \textbf{0.6} \\
    \midrule
    tagging      & VoxelKeypointFusion           & 89.1          & \hspace{6pt} 84.2          & 92.4          & 67.1          & \hspace{6pt} 84.3          & 94.3          & 8.0             & 93.1           & \hspace{6pt} & 207 & 379                         \\
                 & RapidPoseTriangulation        & \textbf{97.5} & \hspace{6pt} \textbf{92.9} & \textbf{99.4} & \textbf{47.9} & \hspace{6pt} \textbf{90.0} & \textbf{100}  & \textbf{4.1}    & \textbf{97.9}  & \hspace{6pt} 11.0 & \textbf{26.5} & \textbf{0.3} \\
    \midrule
    fencing      & VoxelKeypointFusion           & \textbf{100}  & \hspace{6pt} 97.7          & \textbf{100}  & 37.5          & \hspace{6pt} \textbf{100}  & \textbf{100}  & 69.0            & 47.3           & \hspace{6pt} & 335 & 420                         \\
                 & RapidPoseTriangulation        & \textbf{100}  & \hspace{6pt} \textbf{99.6} & \textbf{100}  & \textbf{23.9} & \hspace{6pt} \textbf{100}  & \textbf{100}  & \textbf{50.0}   & \textbf{66.7}  & \hspace{6pt} 12.3 & \textbf{27.9} & \textbf{0.4} \\
    \midrule
    basketball   & VoxelKeypointFusion           & 96.3          & \hspace{6pt} 92.2          & 97.6          & 53.5          & \hspace{6pt} 94.8          & 98.2          & 50.9            & 65.5           & \hspace{6pt} & 212 & 804                         \\
                 & RapidPoseTriangulation        & \textbf{99.6} & \hspace{6pt} \textbf{96.3} & \textbf{100}  & \textbf{40.8} & \hspace{6pt} \textbf{99.7} & \textbf{100}  & \textbf{24.8}   & \textbf{85.8}  & \hspace{6pt} 11.0 & \textbf{27.1} & \textbf{0.4} \\
    \midrule
    volleyball   & VoxelKeypointFusion           & 97.1          & \hspace{6pt} 94.1          & 98.0          & 48.5          & \hspace{6pt} 96.1          & 98.6          & 25.2            & 85.0           & \hspace{6pt} & 449 & 1004                        \\
                 & RapidPoseTriangulation        & \textbf{100}  & \hspace{6pt} \textbf{99.2} & \textbf{100}  & \textbf{27.6} & \hspace{6pt} \textbf{100} & \textbf{100}  & \textbf{5.8}    & \textbf{97.0}  & \hspace{6pt} 18.0 & \textbf{56.4} & \textbf{1.1} \\
    \midrule
    badminton    & VoxelKeypointFusion           & 99.8          & \hspace{6pt} 99.0          & \textbf{100}  & 36.5          & \hspace{6pt} \textbf{100}  & \textbf{100}  & 16.2            & 91.2           & \hspace{6pt} & 391 & 846                         \\
                 & RapidPoseTriangulation        & \textbf{99.9} & \hspace{6pt} \textbf{99.5} & \textbf{100}  & \textbf{24.3} & \hspace{6pt} \textbf{100}  & \textbf{100}  & \textbf{0.0}    & \textbf{100}   & \hspace{6pt} 17.5 & \textbf{42.1} & \textbf{0.7} \\
    \bottomrule
  \end{tabular}
  \caption{Transfer to \textit{egohumans}~\cite{khirodkar2023ego}}
  \vspace{-9pt}
  \label{tab:trans_egohumans}
\end{table*}

Because the cameras use fisheye lenses, the images are more strongly distorted than in the other datasets.
Therefore, a different undistortion method is required. 
Since this requires larger changes in the pipeline of most algorithms, only a subset of the better performing algorithms was tested on this dataset.
In some cases the poses can be undistorted after the 2D estimation step, in other cases, the images need to be undistorted instead.
\textit{VoxelPose} and \textit{Faster-VoxelPose}, for which the images are undistorted beforehand, both have problems with a direct transfer to the \textit{legoassemble} subset, but again show much better performance after synthetic training.
The same holds true for \textit{VoxelPose} on the \textit{tennis} subset, \textit{Faster-VoxelPose} on the other hand does not work at all here.
The synthetic training always failed with an out-of-memory error, even at the lowest possible batch-size.
For \textit{mvpose} the number of views was too high, the six cameras from \textit{tsinghua} already took up the complete GPU memory.
\textit{TEMPO}~\cite{choudhury2023tempo} described an evaluation on this dataset, in which they reported an average transfer error of $120\,mm$ (or $36.7\,mm$ after training) on this dataset, but they did not provide further information about the evaluation in the paper or their code.
\textit{VoxelKeypointFusion} proves its generalization capabilities with a very good performance, which is only surpassed by \textit{RapidPoseTriangulation}.

\subsection{About real-time performance}
\vspace{-3pt}

Regarding real-time performance with many cameras, the results show that while the 3D part of
\textit{RapidPoseTriangulation} is very fast and takes at most $1\,ms$, the majority of the time is currently spent on 2D pose estimation.

In comparison to the original implementation of \textit{RTMPose}, the preprocessing pipeline and network interfacing are already optimized.
Some parts of the preprocessing were moved to the GPU, and the image is transferred as \textit{uint8} type to decrease transfer time.
The box cropping and padding were optimized, and for image resizing nearest-neighbor instead of linear interpolation is used for further speedups.
Everything was implemented in \textit{C++} as well.
Cropping and resizing directly on the GPU did not show notable speedups, as transferring the full image counterweighted its benefits.

To further reduce the input latency in real applications, the \textit{Bayer} format, which is the default layout of image sensors in most cameras, is used directly.
Instead of having three RGB channels, this format only has one channel, in which every other pixel has a single red, green, or blue color only.
The conversion into RGB is now integrated into the preprocessing pipeline.
For the dataset evaluation, the RGB images were first converted back to \textit{Bayer} encoding.
Interestingly, this back-and-forth conversion notably improved the \textit{MPJPE} score in many datasets, presumably because the demosaicing algorithm of \textit{OpenCV} is better than the default one of many cameras.
The demosaicing time is already included in the \textit{FPS} values of the previous tables.
In the case of \textit{FullHD} resolution, it is around $0.2\,ms$ per image.
With \textit{4K} inputs, like in the \textit{egohumans} dataset, it increases to a notable $1.4\,ms$ per image, so demosaicing after cropping and resizing, or directly using \textit{Bayer} images as network input, might be future optimization possibilities.

\vspace{3pt}
In real-time applications, it is also important to weigh algorithmic accuracy versus its latency.
Because persons normally move around, estimations with a high latency result in a larger error to the real position of the human.
For example, a person moving with $2\,m/s$ would move $20\,mm$ in only $10\,ms$, which is already more than the average \textit{MPJPE} difference between the better-performing algorithms from Table~\ref{tab:trans_avg6}.
Algorithmic speed therefore can bring a significant real-time-position accuracy improvement.

\vspace{3pt}
Another important latency factor in real-time systems is the transfer time of the image from the camera to the computer, especially if multiple cameras feed one host.
Transferring a single \textit{FullHD} image with $1920\times1080\times1$ pixels in \textit{Bayer} encoding over a $1\,Gb/s$
network already takes around $17\,ms$.
And this has to be stacked for every further camera. While compressing the image could reduce this time, it would require extra processing time and reduce the image quality. 
A different solution could be adding a small computer to each camera, thus creating a smart-edge device, which estimates the poses for only one image and sends only a few keypoint coordinates to the main computer.
Because only a single image is processed per time-step, this also would allow less powerful hardware.
A \textit{Raspberry Pi} with the \textit{AI-Hat} could already be sufficient.
If the price is not a concern, each camera could be equipped with one standard GPU though.
The RTX-3090 used before takes about $1\,ms$ per image and $1\,ms$ per person to estimate their 2D pose.
With an RTX-4080 this was even $45\%$ faster, and with an RTX-5090 (which has about $2.5\times$ more FLOPS than the RTX-3090), it should be possible to achieve full image-to-pose 3D prediction in only one millisecond.


\section{Whole-body Estimation}
\label{sec:wholebody}
\vspace{-3pt}

\begin{table*}[!htbp]
  \vspace{3pt}
  \fontsize{7pt}{7pt}\selectfont
  \centering
  \begin{tabular}{@{}|l|c|cc|c|cc|c|c|c|@{}}
    \toprule
    Method \hspace{2pt}                   & {\,}PCP{\,}   & \multicolumn{2}{c|}{PCK@100/500}           & {\,}MPJPE (All|Body|Face|Hands){\,}                           & \multicolumn{2}{c|}{Recall@100/500}              & {\,}Invalid{\,} & {\,}FPS{\,}   \\
    \midrule
    VoxelKeypointFusion~\cite{voxkeyfuse} & 99.7          & \hspace{6pt} \textbf{96.8} & 99.8          & \textbf{40.5} | 37.3          | \textbf{38.3} | \textbf{40.7} & \hspace{6pt} 98.5                 & \textbf{100} & 0.5             & 4.9           \\
    RapidPoseTriangulation                & \textbf{99.8} & \hspace{6pt} 88.3          & \textbf{100}  & 45.7          | \textbf{29.1} | 54.3          | 41.8          & \hspace{6pt} \textbf{100}         & \textbf{100} & \textbf{0}      & \textbf{82.6} \\
    \bottomrule
  \end{tabular}
  \caption{Whole-body estimation on \textit{h3wb}.}
  \label{tab:h3wb}
  \vspace{-6pt}
\end{table*}

As already stated in \textit{VoxelKeypointFusion}~\cite{voxkeyfuse}, a significant benefit of algorithmic approaches over the learning-based ones, is their ability to handle input keypoints that were not part of the training dataset, such as \textit{whole-body} joints.

To evaluate the performance, the \textit{h3wb} dataset~\cite{Zhu_2023_ICCV} was used, which extended a part of the \textit{human36m} dataset with whole-body keypoints (17 body, 6 foot, 68 face, 42 hand).
For simplification of the labeling process, only frames with good visibility of the actor were used.
Therefore, the dataset is somewhat simpler than the original one (the default model of \textit{RapidPoseTriangulation} reaches a \textit{MPJPE} of $23.7\,mm$, and the authors estimated around $17\,mm$ labeling error). 
The whole-body results can be found in Table~\ref{tab:h3wb}.
Since this dataset is normally used for single-view 2D-to-3D pose lifting, no other comparable results were found.

One problem of \textit{VoxelKeypointFusion} are the artifacts of the voxelization process, which, due to the discretization, push some keypoints closer together than they really are. 
This was especially visible at the finger keypoints, which were often agglomerated together instead of showing distinct finger lines. 
\textit{RapidPoseTriangulation} does not have this problem anymore, since it directly triangulates in continuous coordinates. 
On the other hand, smaller errors in the 2D keypoint prediction can lead to larger errors in 3D.
For example, if a keypoint is visible from three views, \textit{VoxelKeypointFusion} merges them all at once using heatmap-beams.
So if there are small 2D position errors, the beams of all three views still partially overlap.
\textit{RapidPoseTriangulation} on the other hand triangulates each pair of views separately, and merges them afterward, but the errors do not necessarily cancel themselves out again.
This is more severe if the keypoints can only be seen in very few views, such as in this dataset, since it is not possible to detect and drop outliers before the merging step.

A large difference can be seen in the inference speed, especially in the 3D part.
\textit{VoxelKeypointFusion} took on average $122\,ms$ for the whole-body prediction.
\textit{RapidPoseTriangulation} on the other hand only took $0.1\,ms$ for the whole-body prediction, which is around $1100\times$ faster.


\vspace{-3pt}
\section{Conclusion}
\label{sec:conclusion}
\vspace{-3pt}

This work presented \textit{RapidPoseTriangulation}, a new algorithm for 3D human pose triangulation. 
Its basic idea is to triangulate pairs of 2D poses from different views separately, and group and merge them to the final poses directly in 3D space.
This relatively simple concept makes it very efficient, and much faster than any previous algorithm. 

In a broad evaluation of the algorithm on several datasets, it showed state-of-the-art performance.
It scales well with an increased number of cameras and is especially suited for real-time applications.
Besides that, the results also show that a relatively simple algebraic approach still can outperform many recent fully differentiable or end-to-end learned multi-view methods.
This suggests that the current trend of increasing model complexity does not necessarily translate into better exploitation of the geometric structure.

In summary, it is one of the best and by far the fastest method for multi-view multi-person human pose estimation that is currently available. 
In contrast to most prior works, it is also able to predict whole-body poses, which can be used to implement more advanced follow-up tasks.
By making the source-code publicly available, it is hoped that this work can contribute to the development of more intuitive and safer human-computer or human-robot interactions.


\newpage

{
    \fontsize{8.8}{10.8}\selectfont
    \bibliographystyle{ieeenat_fullname}
    \bibliography{main.bib}

@String(CVPR= {IEEE Conf. Comput. Vis. Pattern Recog.})

@String(ICCV= {Int. Conf. Comput. Vis.})

@String(ECCV= {Eur. Conf. Comput. Vis.})

@String(TOG= {ACM Trans. Graph.})

@String(CVPR  = {CVPR})

@String(ICCV  = {ICCV})

@String(ECCV  = {ECCV})

@String(TOG   = {ACM TOG})

@article{
  voxkeyfuse,
  title={{VoxelKeypointFusion: Generalizable Multi-View Multi-Person Pose Estimation}},
  author={Bermuth, Daniel and Poeppel, Alexander and Reif, Wolfgang},
  journal={arXiv preprint arXiv:2410.18723},
  year={2024}
}

@article{
  simdepose,
  title={{SimpleDepthPose: Fast and Reliable Human Pose Estimation with RGBD-Images}},
  author={Bermuth, Daniel and Poeppel, Alexander and Reif, Wolfgang},
  journal={arXiv preprint arXiv:2501.18478},
  year={2025}
}

@article{jiang2023rtmpose,
  title={{RTMPose: Real-Time Multi-Person Pose Estimation based on MMPose}},
  author={Jiang, Tao and Lu, Peng and Zhang, Li and Ma, Ningsheng and Han, Rui and Lyu, Chengqi and Li, Yining and Chen, Kai},
  journal={arXiv preprint arXiv:2303.07399},
  year={2023}
}

@inproceedings{belagiannis20143d,
  title={{3D pictorial structures for multiple human pose estimation}},
  author={Belagiannis, Vasileios and Amin, Sikandar and Andriluka, Mykhaylo and Schiele, Bernt and Navab, Nassir and Ilic, Slobodan},
  booktitle={Proceedings of the IEEE conference on computer vision and pattern recognition},
  pages={1669--1676},
  year={2014}
}

@article{h36m_pami,
author = {Ionescu, Catalin and Papava, Dragos and Olaru, Vlad and Sminchisescu,  Cristian},
title = {{Human3.6M: Large Scale Datasets and Predictive Methods for 3D Human Sensing in Natural Environments}},
journal = {IEEE Transactions on Pattern Analysis and Machine Intelligence},
publisher = {IEEE Computer Society},
volume = {36},
number = {7},
pages = {1325-1339},
month = {jul},
year = {2014}
}

@article{srivastav2018mvor,
  title={{MVOR: A multi-view RGB-D operating room dataset for 2D and 3D human pose estimation}},
  author={Srivastav, Vinkle and Issenhuth, Thibaut and Kadkhodamohammadi, Abdolrahim and de Mathelin, Michel and Gangi, Afshin and Padoy, Nicolas},
  journal={arXiv preprint arXiv:1808.08180},
  year={2018}
}

@inproceedings{joo2015panoptic,
  title={{Panoptic studio: A massively multiview system for social motion capture}},
  author={Joo, Hanbyul and Liu, Hao and Tan, Lei and Gui, Lin and Nabbe, Bart and Matthews, Iain and Kanade, Takeo and Nobuhara, Shohei and Sheikh, Yaser},
  booktitle={Proceedings of the IEEE International Conference on Computer Vision},
  pages={3334--3342},
  year={2015}
}

@InProceedings{Zhu_2023_ICCV,
    author    = {Zhu, Yue and Samet, Nermin and Picard, David},
    title     = {{H3WB: Human3.6M 3D WholeBody Dataset and Benchmark}},
    booktitle = {Proceedings of the IEEE/CVF International Conference on Computer Vision (ICCV)},
    month     = {October},
    year      = {2023},
    pages     = {20166-20177}
}

@InProceedings{20204DAssociation,
  author = {Zhang, Yuxiang and An, Liang and Yu, Tao and Li, xiu and Li, Kun and Liu, Yebin},
  title = {{4D Association Graph for Realtime Multi-person Motion Capture Using Multiple Video Cameras}},
  booktitle = {IEEE International Conference on Computer Vision and Pattern Recognition, (CVPR)},
  year={2020},
}

@inproceedings{khirodkar2023ego,
  title={{Ego-Humans: An Ego-Centric 3D Multi-Human Benchmark}},
  author={Khirodkar, Rawal and Bansal, Aayush and Ma, Lingni and Newcombe, Richard and Vo, Minh and Kitani, Kris},
  booktitle={Proceedings of the IEEE/CVF International Conference on Computer Vision},
  pages={19807--19819},
  year={2023}
}

@inproceedings{voxelpose,
    author={Tu, Hanyue and Wang, Chunyu and Zeng, Wenjun},
    title={{VoxelPose: Towards Multi-Camera 3D Human Pose Estimation in Wild Environment}},
    booktitle = {European Conference on Computer Vision (ECCV)},
    year = {2020}
}

@inproceedings{iskakov2019learnable,
  title={{Learnable triangulation of human pose}},
  author={Iskakov, Karim and Burkov, Egor and Lempitsky, Victor and Malkov, Yury},
  booktitle={Proceedings of the IEEE/CVF international conference on computer vision},
  pages={7718--7727},
  year={2019}
}

@inproceedings{fastervoxelpose,
    author={Ye, Hang and Zhu, Wentao and Wang, Chunyu and Wu, Rujie and Wang, Yizhou},
    title={{Faster VoxelPose: Real-time 3D Human Pose Estimation by Orthographic Projection}},
    booktitle = {European Conference on Computer Vision (ECCV)},
    year = {2022}
}

@inproceedings{choudhury2023tempo,
  title={{TEMPO: Efficient multi-view pose estimation, tracking, and forecasting}},
  author={Choudhury, Rohan and Kitani, Kris M and Jeni, L{\'a}szl{\'o} A},
  booktitle={Proceedings of the IEEE/CVF International Conference on Computer Vision},
  pages={14750--14760},
  year={2023}
}

@inproceedings{reddy2021tessetrack,
  title={{Tessetrack: End-to-end learnable multi-person articulated 3d pose tracking}},
  author={Reddy, N Dinesh and Guigues, Laurent and Pishchulin, Leonid and Eledath, Jayan and Narasimhan, Srinivasa G},
  booktitle={Proceedings of the IEEE/CVF conference on computer vision and pattern recognition},
  pages={15190--15200},
  year={2021}
}

@inproceedings{wu2021graph,
  title={{Graph-based 3d multi-person pose estimation using multi-view images}},
  author={Wu, Size and Jin, Sheng and Liu, Wentao and Bai, Lei and Qian, Chen and Liu, Dong and Ouyang, Wanli},
  booktitle={ICCV},
  year={2021}
}

@article{wang2021mvp,
  title={{Direct Multi-view Multi-person 3D Human Pose Estimation}},
  author={Tao Wang and Jianfeng Zhang and Yujun Cai and Shuicheng Yan and Jiashi Feng},
  journal={Advances in Neural Information Processing Systems},
  year={2021}
}

@InProceedings{Srivastav_2024_CVPR,
    author    = {Srivastav, Vinkle and Chen, Keqi and Padoy, Nicolas},
    title     = {{SelfPose3d: Self-Supervised Multi-Person Multi-View 3d Pose Estimation}},
    booktitle = {Proceedings of the IEEE/CVF Conference on Computer Vision and Pattern Recognition (CVPR)},
    month     = {June},
    year      = {2024},
    pages     = {2502-2512}
}

@inproceedings{dong2019fast,
  title={{Fast and Robust Multi-Person 3D Pose Estimation from Multiple Views}},
  author={Dong, Junting and Jiang, Wen and Huang, Qixing and Bao, Hujun and Zhou, Xiaowei},
  journal={CVPR},
  year={2019}
}

@inproceedings{tanke2019iterative,
  title={{Iterative Greedy Matching for 3D Human Pose Tracking from Multiple Views}},
  author={Tanke, Julian and Gall, Juergen},
  booktitle={German Conference on Pattern Recognition},
  year={2019}
}

@inproceedings{chu2021part,
  title={{Part-aware measurement for robust multi-view multi-human 3d pose estimation and tracking}},
  author={Chu, Hau and Lee, Jia-Hong and Lee, Yao-Chih and Hsu, Ching-Hsien and Li, Jia-Da and Chen, Chu-Song},
  booktitle={Proceedings of the IEEE/CVF Conference on Computer Vision and Pattern Recognition},
  pages={1472--1481},
  year={2021}
}

@inproceedings{zhou2022quickpose,
  title={Quickpose: Real-time multi-view multi-person pose estimation in crowded scenes},
  author={Zhou, Zhize and Shuai, Qing and Wang, Yize and Fang, Qi and Ji, Xiaopeng and Li, Fashuai and Bao, Hujun and Zhou, Xiaowei},
  booktitle={ACM SIGGRAPH 2022 Conference Proceedings},
  pages={1--9},
  year={2022}
}

@InProceedings{Chen_2020_CVPR,
author = {Chen, Long and Ai, Haizhou and Chen, Rui and Zhuang, Zijie and Liu, Shuang},
title = {Cross-View Tracking for Multi-Human 3D Pose Estimation at Over 100 FPS},
booktitle = {The IEEE/CVF Conference on Computer Vision and Pattern Recognition (CVPR)},
month = {June},
year = {2020}
}

@article{shuai2023reconstructing,
  title={{Reconstructing Close Human Interactions from Multiple Views}},
  author={Shuai, Qing and Yu, Zhiyuan and Zhou, Zhize and Fan, Lixin and Yang, Haijun and Yang, Can and Zhou, Xiaowei},
  journal={ACM Transactions on Graphics (TOG)},
  volume={42},
  number={6},
  pages={1--14},
  year={2023},
  publisher={ACM New York, NY, USA}
}

@inproceedings{bridgeman2019multi,
  title={{Multi-person 3d pose estimation and tracking in sports}},
  author={Bridgeman, Lewis and Volino, Marco and Guillemaut, Jean-Yves and Hilton, Adrian},
  booktitle={Proceedings of the IEEE/CVF conference on computer vision and pattern recognition workshops},
  pages={0--0},
  year={2019}
}

@inproceedings{fieraru2020three,
  title={{Three-dimensional reconstruction of human interactions}},
  author={Fieraru, Mihai and Zanfir, Mihai and Oneata, Elisabeta and Popa, Alin-Ionut and Olaru, Vlad and Sminchisescu, Cristian},
  booktitle={Proceedings of the IEEE/CVF Conference on Computer Vision and Pattern Recognition},
  pages={7214--7223},
  year={2020}
}

@inproceedings{xiao2018simple,
  title={{Simple baselines for human pose estimation and tracking}},
  author={Xiao, Bin and Wu, Haiping and Wei, Yichen},
  booktitle={Proceedings of the European conference on computer vision (ECCV)},
  pages={466--481},
  year={2018}
}

@inproceedings{sun2019deep,
  title={{Deep high-resolution representation learning for human pose estimation}},
  author={Sun, Ke and Xiao, Bin and Liu, Dong and Wang, Jingdong},
  booktitle={Proceedings of the IEEE/CVF conference on computer vision and pattern recognition},
  pages={5693--5703},
  year={2019}
}

@inproceedings{chen2018cascaded,
  title={{Cascaded pyramid network for multi-person pose estimation}},
  author={Chen, Yilun and Wang, Zhicheng and Peng, Yuxiang and Zhang, Zhiqiang and Yu, Gang and Sun, Jian},
  booktitle={Proceedings of the IEEE conference on computer vision and pattern recognition},
  pages={7103--7112},
  year={2018}
}

@article{cao2019openpose,
  title={{Openpose: Realtime multi-person 2d pose estimation using part affinity fields}},
  author={Cao, Zhe and Hidalgo, Gines and Simon, Tomas and Wei, Shih-En and Sheikh, Yaser},
  journal={IEEE transactions on pattern analysis and machine intelligence},
  volume={43},
  number={1},
  pages={172--186},
  year={2019},
  publisher={IEEE}
}

@inproceedings{lin2014microsoft,
  title={{Microsoft COCO: Common Objects in Context}},
  author={Lin, Tsung-Yi and Maire, Michael and Belongie, Serge and Hays, James and Perona, Pietro and Ramanan, Deva and Doll{\'a}r, Piotr and Zitnick, C Lawrence},
  booktitle={European conference on computer vision},
  pages={740--755},
  year={2014},
  organization={Springer}
}
}

\newpage

\clearpage
\appendix


\section{Ablation Studies}
\vspace{-3pt}

The following section contains the ablation studies that were left out of the main paper for space reasons.

\vspace{-3pt}
\subsection{Number of cameras}
\vspace{-3pt}

For completeness, an experiment with \textit{panoptic} using $3$, $5$, $7$, $10$, or $31$ cameras was conducted.
As expected, the results show that more cameras lead to better results.
The frame-rate also slows down, with the \textit{2D}-part having the most impact, and the \textit{3D}-part having a slight exponential increase, even though its impact is still very small.

For real-world applications, this experiment is not very meaningful though, since the required number of cameras is highly dependent on the targeted environment.
Depending on the setup, $3$ cameras can be enough or much too few, as in \textit{campus} versus \textit{mvor}.
\textit{VP} and \textit{F-VP} both conducted an experiment for $3$ cameras on \textit{panoptic} in which they had only very small performance drops, but their performance on \textit{campus} and \textit{mvor} showed a large drop, even though they also have $3$ cameras.
For a real-world application, it is therefore better to first select a similar dataset and then evaluate the impact of the number of cameras and their best positioning.
Since it is not possible to evaluate all combinations of datasets, cameras, and positions, the recommendation for developers is to use the provided source-code and run a personalized evaluation on the most similar dataset.

\begin{table*}[hbp]
    \fontsize{7pt}{7pt}\selectfont
    \centering
    \begin{tabular}{@{}|l|c|cc|c|cc|c|c|c|c|c|@{}}
      \toprule
      Number of cameras     & {\,}PCP{\,}   & \multicolumn{2}{c|}{PCK@100/500}           & {\,}MPJPE{\,} & \multicolumn{2}{c|}{Recall@100/500}        & {\,}Invalid{\,} & {}{\,}F1{\,}{} & Time-3D & {\,}FPS{\,}   \\
      \midrule
      3                     & 90.4          & \hspace{6pt} 83.7          & 93.5          & 58.2          & \hspace{6pt} 85.2          & 94.5          & 1.1             & 96.7           &   0.1   & 95.8          \\
      5                     & 99.1          & \hspace{6pt} 96.5          & 99.8          & 30.6          & \hspace{6pt} 99.3          & 99.8          & 1.8             & 99.0           &   0.2   & 57.5          \\
      7                     & 99.5          & \hspace{6pt} 97.2          & 99.9          & 27.9          & \hspace{6pt} 99.9          & 99.9          & 3.7             & 98.1           &   0.3   & 40.9          \\
      10                    & 99.7          & \hspace{6pt} 98.0          & 99.9          & 23.9          & \hspace{6pt} 99.9          & 99.9          & 3.0             & 98.4           &   0.5   & 29.6          \\
      31                    & 99.8          & \hspace{6pt} 98.7          & 99.9          & 19.3          & \hspace{6pt} 99.9          & 99.9          & 2.8             & 98.6           &   5.0   & 9.1           \\
      \bottomrule
    \end{tabular}
    \caption{Using different numbers of cameras on \textit{panoptic}.}
    \label{tab:res_cam_abls}
\end{table*}

\begin{table*}[hbp]
    \fontsize{7pt}{7pt}\selectfont
    \centering
    \begin{tabular}{@{}|l|c|c|cc|c|cc|c|c|c|c|c|@{}}
      \toprule
      Ablation     & Dataset   & {\,}PCP{\,}   & \multicolumn{2}{c|}{PCK@100/500}           & {\,}MPJPE{\,} & \multicolumn{2}{c|}{Recall@100/500}        & {\,}Invalid{\,} & {}{\,}F1{\,}{} & Time-3D \\
      \midrule
      default                     & shelf      & 99.2  & \hspace{6pt} 94.4 & 100  & 47.5 & \hspace{6pt} 98.7 & 100  & 41.3   & 74.0  &   0.1 \\
      score without 2D confidence & shelf      & 99.2  & \hspace{6pt} 94.4 & 100  & 47.7 & \hspace{6pt} 98.7 & 100  & 41.3   & 74.0  &   0.1 \\
      no pair pre-filtering       & shelf      & 99.2  & \hspace{6pt} 94.4 & 100  & 47.6 & \hspace{6pt} 98.7 & 100  & 40.4   & 74.6  &   0.2 \\
      keep topk outliers          & shelf      & 98.9  & \hspace{6pt} 93.8 & 100  & 49.8 & \hspace{6pt} 97.9 & 100  & 41.3   & 74.0  &   0.1 \\
      keep topk+distance outliers & shelf      & 98.9  & \hspace{6pt} 93.7 & 100  & 49.9 & \hspace{6pt} 97.9 & 100  & 41.3   & 74.0  &   0.1 \\
      min\_match\_score=0.8       & shelf      & 99.2  & \hspace{6pt} 94.5 & 100  & 47.7 & \hspace{6pt} 97.9 & 100  & 46.9   & 69.3  &   0.1 \\
      \midrule
      default                     & panoptic   & 99.1  & \hspace{6pt} 96.5 & 99.8 & 30.6 & \hspace{6pt} 99.3 & 99.8 & 1.8    & 99.0  &   0.2 \\
      no postprocessing steps     & panoptic   & 99.0  & \hspace{6pt} 96.3 & 99.8 & 31.6 & \hspace{6pt} 99.2 & 99.8 & 1.8    & 99.0  &   0.1 \\
      \midrule
      default                     & badminton  & 99.9  & \hspace{6pt} 99.5 & 100  & 24.3 & \hspace{6pt} 100  & 100  & 0.0    & 100   &   0.7 \\
      only triangulation error    & badminton  & 99.9  & \hspace{6pt} 99.6 & 100  & 24.6 & \hspace{6pt} 100  & 100  & 0.0    & 100   &   0.5 \\
      only reprojection error     & badminton  & 100   & \hspace{6pt} 99.5 & 100  & 24.4 & \hspace{6pt} 100  & 100  & 0.0    & 100   &   0.7 \\
      min\_group\_size=1          & badminton  & 99.4  & \hspace{6pt} 98.9 & 99.8 & 27.1 & \hspace{6pt} 99.3 & 100  & 39.0   & 75.8  &   1.0 \\
      \midrule
      default                     & tagging    & 97.5  & \hspace{6pt} 92.9 & 99.4  & 47.9 & \hspace{6pt} 90.0  & 100  & 4.1    & 97.9  &   0.3 \\
      no tracking                 & tagging    & 97.5  & \hspace{6pt} 93.0 & 99.3  & 46.7 & \hspace{6pt} 91.8  & 99.8 & 3.9    & 97.9  &   0.3 \\
      \bottomrule
    \end{tabular}
    \caption{Different algorithmic ablations.}
    \label{tab:res_abls}
\end{table*}

\vspace{-3pt}
\subsection{Algorithmic ablations}
\vspace{-3pt}

In general, the algorithm is very robust against small changes, and only a few of them have a notable impact on the results.
Ignoring the 2D confidence scores instead to integrating them into the 3D keypoint score, results in a slightly higher \textit{MPJPE}. 
Removing the pair pre-filtering step, which removes pairs of poses before the triangulation, has minimal impact on most scores, but notably increases the triangulation time, since more pairs have to be calculated.
The two outlier removal steps, before merging the proposals of one group to a single person, notably improve the accuracy. 
Lowering the error threshold, which filters poor triangulations before merging, slightly increases \textit{MPJPE} and the invalid count.
The postprocessing steps of dropping clearly invalid persons or replacing clearly wrong joints with their neighbors also improves the \textit{MPJPE} and reduces the invalid person count in some datasets.
Using only the triangulation error or only the reprojection error for the 3D keypoint score calculation, instead of a combination of both, has a slightly negative impact on the \textit{MPJPE}.
Using only the triangulation score is slightly faster though, because the more complex reprojection can be skipped.
Especially in datasets with more cameras, it is recommended to increase the minimal required group size.
This results in dropping invalid 3D proposals which do not originate from the same person, but have a similar pose in two or more views.
The higher the number of cameras the higher the chance for it.
Since they are from different persons, they are triangulated to some mathematically valid positions somewhere in the 3D space.
Proposals from the same person on the other hand start to build clusters around their real position, by which they can be recognized as such.
The tracking step has only a minimal impact on most datasets, since it does not smooth the movement trajectories.
At fast movements it might slightly reduce the position accuracy because it can slow down some position changes.
On the other hand it can prevent physically impossible fast jitter movements.
In case a person is not visible for a few frames, it can also prevent the person from being lost completely, since their last position is returned then, but it will also keep persons moving out of the room alive for a few extra frames.
Therefore tracking is considered as an optional extension.

\section{Further Remarks}
\vspace{-3pt}

The following section contains some further remarks about the algorithmic performance.

\definecolor{deepred}{rgb}{0.6,0,0}
\definecolor{deepgreen}{rgb}{0,0.5,0}

\begin{table*}[hbp]
  \fontsize{7pt}{7pt}\selectfont
  \centering
  \vspace{-6pt}
  \begin{tabular}{@{}|l|l|c|cc|c|cc|c|c|c|@{}}
    \toprule
    Method \hspace{44pt} & Dataset         & {\,}PCP{\,}   & \multicolumn{2}{c|}{PCK@100/500}           & {\,}MPJPE{\,} & \multicolumn{2}{c|}{Recall@100/500}        & {\,}Invalid{\,} & {}{\,}F1{\,}{} \\
    \midrule

    mv3dpose                      & shelf  & 92.2 \textcolor{deepred}{(-4.9)} & \hspace{6pt} 88.2 \textcolor{deepred}{(-3.4)} & 98.4 \textcolor{deepred}{(-4.8)}         & 51.1 \textcolor{deepgreen}{(-4.7)} & \hspace{6pt} 90.6 \textcolor{deepred}{(-4.2)}          & 93.7 \textcolor{deepred}{(-4.8)}        & \textbf{40.6} \textcolor{deepgreen}{(-3.7)}           & 72.7 \textcolor{deepgreen}{(+1.5)}            \\
    PartAwarePose                 & shelf  & 98.3 (0.0)   & \hspace{6pt} 93.5 \textcolor{deepgreen}{(+0.8)} & 99.0 (0.0)         & 48.1 \textcolor{deepgreen}{(-3.3)} & \hspace{6pt} 97.5 \textcolor{deepred}{(-1.0)}          & 99.2 (0.0)        & 49.4 \textcolor{deepred}{(+2.0)}           & 67.0 \textcolor{deepred}{(-1.7)}            \\
    RapidPoseTriangulation        & shelf  & \textbf{99.2} & \hspace{6pt} \textbf{94.4} & \textbf{100} & \textbf{47.5} & \hspace{6pt} \textbf{98.7}    & \textbf{100} & 41.3   & \textbf{74.0}  \\
    \midrule
    mv3dpose                      & campus & 87.1 \textcolor{deepgreen}{(+3.0)} & \hspace{6pt} 71.4 \textcolor{deepgreen}{(+7.0)} & 91.4 \textcolor{deepred}{(-2.0)}         & 75.7 \textcolor{deepgreen}{(-59)} & \hspace{6pt} 81.6 \textcolor{deepgreen}{(+19)}          & 91.8 \textcolor{deepred}{(-3.1)}        & 17.5 \textcolor{deepred}{(+7.4)}         & 86.9 \textcolor{deepred}{(-5.5)}          \\
    PartAwarePose                 & campus & 94.2 \textcolor{deepgreen}{(+1.0)} & \hspace{6pt} 78.4 (0.0) & 98.9 (0.0)         & \textbf{74.6} \textcolor{deepgreen}{(-0.1)} & \hspace{6pt} 92.0 \textcolor{deepred}{(-1.9)}          & 98.9 (0.0)        & 17.1 \textcolor{deepgreen}{(-5.6)}         & 90.2 \textcolor{deepgreen}{(+3.4)}          \\
    RapidPoseTriangulation        & campus & \textbf{95.2} & \hspace{6pt} \textbf{79.9} & \textbf{100}  & 75.2          & \hspace{6pt} \textbf{93.9} & \textbf{100} & \textbf{15.9}            & \textbf{91.4}          \\
    \bottomrule
  \end{tabular}
  \caption{Replacing the original 2D pose inputs with those of \textit{RTMPose}, and the change to the original results.}
  \label{tab:switch_detector}
\end{table*}

\subsection{Computation Comparison}

The computation efficiency comparison was made as fair as technically possible.
Results taken from \cite{voxkeyfuse} in Table~\ref{tab:trans_h36m} onwards used either RTX3090 (same as here), or GTX1080 (\textit{mvpose, mv3dpose, PartAwarePose}) because they did not work on newer GPUs.
In Table~\ref{tab:times}, the methods \textit{mvpose, VoxelPose, Faster-VoxelPose, PlaneSweepPose, VoxelKeypointFusion} use a GPU, the rest is on CPU.
The fastest three in the table~([4,5,26]) are closed source, so comparison on the same hardware is not possible.
[4]~used an \mbox{Intel-i7~$3.2\,GHz$}, with parallelization.
[26] and \textit{PartAwarePose} used a $3.7\,GHz$~CPU, [5] a $2.2\,GHz$~CPU, the AMD7900X used here has $4.7\,GHz$, all those works used a single core.
\textit{RapidPoseTriangulation} is at least $30\times$ faster, which is much more than the $GHz$ difference, so the main speedup certainly is from the algorithm itself.

\subsection{Different 2D Estimators}

The referenced previous works use a wide variety of 2D detectors, and all results from Table~\ref{tab:trans_h36m} onwards are with the originally used detectors.
\textit{VoxelPose}, \textit{PRGnet}, \textit{MvP}, \textit{SelfPose3d} all use \textit{ResNet}~\cite{xiao2018simple}, \textit{Faster-VP} and \textit{TEMPO} switched to the newer \textit{HRNet}~\cite{sun2019deep}.
\textit{mvpose} uses \textit{CPN}~\cite{chen2018cascaded}, \textit{mv3dpose} uses \textit{OpenPose}~\cite{cao2019openpose}, \textit{PartAwarePose} uses \textit{HRNet}~\cite{sun2019deep}.
\textit{VoxelKeypointFusion} uses \textit{RTMPose}~\cite{jiang2023rtmpose}, same as this work.
In general, all previous works that switched their pose detector to newer models did not investigate this impact on performance.
Speaking from experience, it is reasonable that this is not a standard practice in the field, because it takes a lot of effort to adjust often poorly documented code to inject different poses into it.
For this reason, this work only conducted a small experiment with \textit{mv3dpose} and \textit{PartAwarePose} using exactly the same 2D poses as \textit{RapidPoseTriangulation}, instead of the ones from their original \textit{OpenPose} and \mbox{\textit{HRNet}} pipelines~(Table~\ref{tab:switch_detector}).
For \textit{mv3dpose} the results show improvements the localization accuracy (\textit{MPJPE}), but a notable drop in person detection (\textit{Recall@500}). 
The results of \textit{PartAwarePose} show only marginal changes, mainly a slightly better \textit{MPJPE} (mostly caused by better hips, while many joints even got worse).
All scores are still clearly behind \textit{RapidPoseTriangulation}, so the detector change is not the main reason for the better performance.

\subsection{Additional Reasoning}

The (voxel-based) learned methods all have generalization problems, they work well on the same dataset (\textit{panoptic}), but show notable loss in setup switches.
The synthetic training reduced this somewhat in many cases, but not to the full extend.
They also do not allow using arbitrary keypoints, but require matching training datasets instead.
This was already evaluated in more detail in~\cite{voxkeyfuse}.
And as \textit{VoxelKeypointFusion} has shown, the problems do not come from the voxel representation itself.
But even if those problems could be solved, those methods are rather slow (as shown in Table~\ref{tab:times}), which makes them less suitable for real-time applications, and, as \textit{VoxelKeypointFusion} has shown, adding
a lot more keypoints makes them even slower.

The algebraic methods often fail in person association.
For example in Table\,2+3 from~\cite{voxkeyfuse} \textit{mv3dpose} and \textit{PartAwarePose} are not able to detect all persons (Recall@500), and as shown here in Table~\ref{tab:switch_detector}, this is not due to the 2D detector.
The same problem can be seen in Table\,5 from \cite{voxkeyfuse} or Table~\ref{tab:trans_tsinghua}.
The better performance of \textit{RapidPoseTriangulation} in this regard can be explained by the different matching strategy.
Unlike the others, \textit{RPT} triangulates all possible view-pairs to 3D poses, and then drops those with large errors (steps 1-8).
So if a person is seen in only two images it is very likely to be reconstructed.
This makes \textit{RapidPoseTriangulation} so reliable, which is why this approach was chosen.
Grouping all those 3D proposals to persons (step 9) is very robust as well, because it is highly unlikely that two persons have very similar 3D poses while standing at basically the same spot.
Because there are no projection caused similarities this is more reliable than 2D pose matching across views, especially in more crowded scenes, which have a higher probability of similar looking persons, and therefore more false matches.
And compared to appearance based matching, it has no problems with similar looking persons (like in \textit{MVOR}), and it is orders of magnitudes faster.
While a larger number of persons also affects the likelihood of wrong 3D proposals generated in \textit{RapidPoseTriangulation}, they can easily be filtered out by setting a higher minimal group size. 

Regarding the joint accuracy (PCK@100), having multiple 3D proposals per person from the different view-pairs allows for a very efficient but robust outlier detection.
At its core it's based on majority voting, which removes very bad proposals, without them influencing the final position, which would not be possible with iterative matching or (weighted) averaging like in other works.

\end{document}